\documentclass[runningheads]{llncs}

\usepackage{eccv}

\usepackage{eccvabbrv}

\usepackage{graphicx}
\usepackage{booktabs}

\usepackage[accsupp]{axessibility}  % Improves PDF readability for those with disabilities.
\usepackage{multirow}
\usepackage{xcolor}
\usepackage{hyperref}

\usepackage{hyperref}

\usepackage{orcidlink}

\begin{document}

% ---------------------------------------------------------------
% TODO REVIEW: Replace with your title
\title{Monocular Avatar Reconstruction via Cascaded Diffusion Priors and UV-Space Differentiable Shading} 

% TODO REVIEW: If the paper title is too long for the running head, you can set
% an abbreviated paper title here. If not, comment out.
\titlerunning{Monocular Avatar Reconstruction}

% TODO FINAL: Replace with your author list. 
% Include the authors' OCRID for the camera-ready version, if at all possible.

\newcommand{\equalcontrib}{\textsuperscript{$\star$}}
\newcommand{\projectlead}{\textsuperscript{\dag}}
\newcommand{\corrauth}{\textsuperscript{\ddag}}

\author{
Hong Li \inst{1,2}\thanks{Equal contribution.
\quad \textsuperscript{\dag}Project leader.
\quad \textsuperscript{\ddag}Corresponding author.}\orcidlink{0000-0002-4240-3073} \and
Minqi Meng\equalcontrib\inst{1}\orcidlink{0009-0005-1567-5646} \and
Yanjun Liang\inst{1}\orcidlink{0009-0008-0568-2254} \and
Chongjie Ye\inst{3}\orcidlink{0000-0002-7123-0220} \and
Houyuan Chen\inst{4}\orcidlink{0009-0005-4693-2326} \and
Weiqing Xiao\inst{5}\orcidlink{0009-0003-2548-0485} \and
Xianda Guo\inst{6}\projectlead\orcidlink{0000-0003-2822-4690} \and
Guojun Lei\inst{7}\orcidlink{0009-0000-6282-218X} \and
Xuhui Liu\inst{8}\orcidlink{0000-0001-6064-3401} \and
Chaojie Yang\inst{1}\orcidlink{0009-0003-8139-0572} \and
Yanlun Peng\inst{2}\orcidlink{0009-0009-6972-7690} \and
Hao Zhao\inst{9}\orcidlink{0000-0001-7903-581X} \and
Baochang Zhang\inst{1}\corrauth\orcidlink{0000-0001-6167-4760}
}

% TODO FINAL: Replace with an abbreviated list of authors.
\authorrunning{Li et al.}

\institute{
\begin{tabular}{cc}
$^{1}$ Beihang University, China & $^{2}$ Great Wall Motors, China \\
$^{3}$ CUHK-Shenzhen, China & $^{4}$ HKUST, China \\
$^{5}$ Nanjing University, China & $^{6}$ Wuhan University, China \\
$^{7}$ Zhejiang University, China & $^{8}$ KAUST, Saudi Arabia \\
\multicolumn{2}{c}{$^{9}$ AIR, Tsinghua University, China}
\end{tabular}
\\
{\footnotesize\textit{Project Page:} \href{https://luh1124.github.io/MARCUS-Avatar-Projectpage/}{\textcolor{magenta}{marcus-avatar.github.io}}}
}

\maketitle

\begin{abstract}
  Reconstructing high-fidelity, relightable 3D avatars from a single in-the-wild image is a challenging ill-posed problem, primarily hindered by the scarcity of high-quality PBR data and the complexity of disentangling illumination from intrinsic materials. In this paper, we present a data-efficient framework that leverages the robust priors of a unified pre-trained diffusion backbone to sequentially address texture completion, delighting, and material decomposition. Unlike existing methods that rely on fragmented pipelines or extensive proprietary datasets, we utilize cascaded Low-Rank Adaptations (LoRAs) to adapt the strong generative prior of the diffusion model for each sub-task in UV space. Specifically, we first employ an Inpainting LoRA to complete missing UV textures caused by occlusion, leveraging the model's semantic understanding to generate semantically and photometrically coherent details. Subsequently, a Light-Homogenization LoRA and a novel Cross-Intrinsic Attention mechanism are introduced to remove baked-in lighting and collaboratively synthesize pixel-aligned PBR maps (Albedo, Normal, Roughness, Specular, and Displacement). To ensure physical plausibility, we impose a UV-space differentiable BRDF shading loss during the decomposition stage, forcing the generative process to adhere to the rendering equation without the artifacts typical of rasterization-based supervision. Extensive experiments demonstrate that our method, trained on fewer than 100 real 3D scans, generates comprehensive, 4K-resolution PBR assets with superior realism and generalization compared to state-of-the-art methods, and all training code and model weights will be released upon acceptance.
  \keywords{3D avatar \and face reconstruction \and Diffusion model \and material estimation \and differentiable rendering}
\end{abstract}

% \vspace{-20pt}

\begin{figure}
  \centering
  \includegraphics[width=\textwidth]{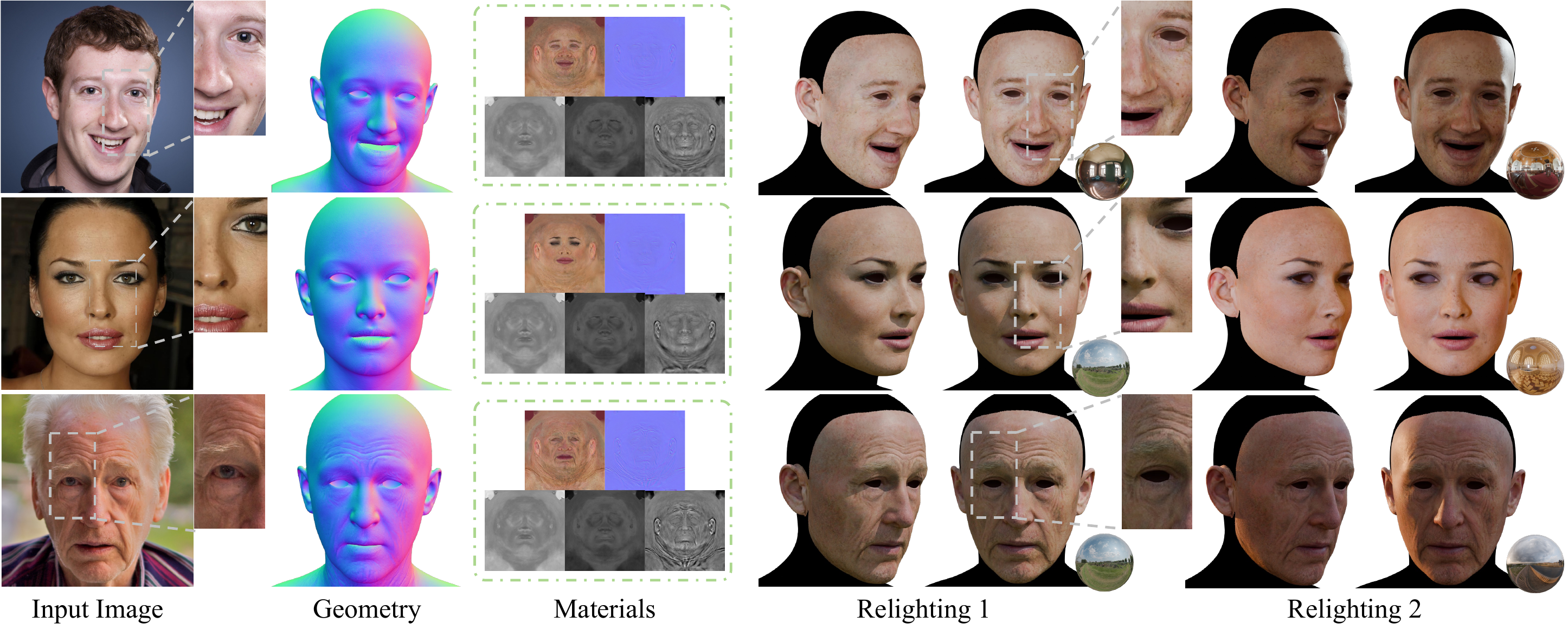}
  \vspace{-20pt}
  \caption{\textbf{High-Fidelity 3D Avatar Reconstruction}. From a single input image, we reconstruct high-fidelity 3D geometry and PBR materials (albedo, normal, packed maps) to enable relightable avatar synthesis. As shown in the novel lighting results (right), our method accurately recovers fine details (e.g., wrinkles, moles) whilst maintaining identity consistency.}
  \label{fig:teaser}
\end{figure}   
% \vspace{-20pt}

\section{Introduction}
\label{sec:intro}

Photorealistic avatars are pivotal to immersive experiences in virtual reality, multimedia, and video games~\cite{wu2025fastavatar, Wang_2025_CVPR}. These personalized digital doubles serve as a crucial interface, bridging physical identities with digital environments. While Light Stage systems~\cite{debevec2012light} have long established the gold standard for acquiring high fidelity avatars, their prohibitive cost and reliance on complex, studio based capture setups severely limit their accessibility for mass adoption. Therefore, monocular 3D face reconstruction, recovering geometry and texture from a single in-the-wild image, remains a fundamental challenge in computer vision. In this paper, we focus on relightable face and head PBR asset reconstruction rather than full-body, dynamic, or accessory-complete avatars.

Traditional approaches rely mainly on 3D Morphable Models (3DMMs)~\cite{blanz2023morphable, paysan20093d}, which represent facial variations within a compact linear subspace. While robust, these linear models inherently struggle to capture high frequency details. To overcome this, recent research has shifted towards neural synthesis paradigms. Early works focused on texture completion~\cite{deng2018uv,gecer2019ganfit,gecer2021ostec, li2024uv, yang2025freeuv}; however, they typically produce textures with baked-in illumination, rendering the assets unsuitable for relighting in physically based rendering (PBR) pipelines.

To achieve disentangled PBR reconstruction, recent methods typically train generative networks (GANs or Diffusion) to predict nonlinear texture bases~\cite{Lattas2022AvatarMe++,lattas2020avatarme,Lattas2023fitme,fitdiff,Papantoniou2023Relightify, dib2024mosar, bai2023ffhq, dai2025high}. However, these approaches face a critical data efficiency and generalization dilemma.
First, high quality PBR training data is scarce: synthetic data often lacks realism, while real Light Stage scans are difficult to scale.
Second, methodological bottlenecks persist: GAN-based methods suffer from mode collapse and over smoothing, while recent attempts to finetune diffusion models on small real datasets (e.g., UltraAvatar~\cite{zhou2024ultravatar}) often degrade the model's prior knowledge, leading to poor generalization on in-the-wild images.
Furthermore, most existing frameworks rely on differentiable rasterization with 2D screen-space supervision. This optimization paradigm is sensitive to occlusions (e.g., hair, glasses) present in input images, often inadvertently "baking" these artifacts into the reconstructed texture.

In this work, we present a novel framework that addresses these challenges by synergizing the robust priors of large-scale generative models with a modern rendering workflow (Fig.~\ref{fig:teaser}). Our core insight is that we can circumvent the need for massive proprietary datasets by effectively augmenting a compact set of high-quality scans via modern rendering engines, incorporating pose and occlusion priors derived from public in-the-wild face datasets. By combining these augmented physical priors with a pre-trained modern diffusion backbone, our method achieves high-fidelity material estimation without requiring extensive real-world data. Crucially, we enforce physically based rendering constraints directly on the estimated materials in UV space. 
% This strategy not only ensures the physical consistency and high quality of the textures but also effectively mitigates the occlusion artifacts commonly observed in previous rasterization-based approaches.
This strategy ensures the physical consistency and high quality of the textures while mitigating the occlusion artifacts commonly observed in previous rasterization-based approaches.

Our pipeline begins with a precise geometric alignment step. To support high-fidelity texture generation, we employ a robust geometry estimator enhanced by DINOv3 features, ensuring accurate UV mapping. With the geometry fixed, our core contribution lies in a unified diffusion-based texture framework adapted via cascaded LoRA modules. This backbone sequentially performs texture inpainting, light homogenization, and intrinsic material estimation. To ensure physical plausibility, we introduce a Cross-Intrinsic Attention mechanism that coordinates distinct material branches (Albedo, Normal, Roughness, Specular, Displacement), ensuring coherent high-frequency details. Trained on fewer than 100 real 3D scans, our model generates comprehensive, 4K-resolution PBR assets that generalize seamlessly to diverse in-the-wild images.

In summary, our key contributions are: 

\begin{itemize} 
    \item We propose a data-efficient framework for high-fidelity 3D avatar reconstruction that achieves state-of-the-art quality using fewer than 100 real 3D scans, effectively solving the data scarcity bottleneck via a Blender-augmented synthetic data pipeline. 
    \item We design a unified diffusion-based texture pipeline featuring a novel Cross-Intrinsic Attention mechanism, which enforces structural coherence and physical plausibility across generated material modalities. 
    \item Leveraging generative priors and UV-space physical constraints, our method effectively avoids the pitfalls of direct 2D supervision, producing 4K-resolution PBR assets robust to occlusions and generalizing well to in-the-wild inputs. 
\end{itemize}

\section{Related Works}
\label{sec:relatedwork}

\subsection{3D Geometry Reconstruction}
% Monocular 3D face geometry reconstruction is a fundamental challenge in computer vision~\cite{wang2021review}. Existing approaches primarily fall into two categories: coefficient regression based on parametric statistical models and methods leveraging non-linear models. The first category predominantly utilizes parametric 3D face models, notably the 3D Morphable Model (3DMM)~\cite{blanz2023morphable, hifi3dface2021tencentailab, wang2022faceverse, FaceKit} and blendshape-driven deformation models~\cite{FLAME}. These methods reconstruct faces by regressing shape, expression, pose, and illumination coefficients from a single 2D image. Recent advances largely leverage deep learning to directly predict 3DMM coefficients in an end-to-end manner~\cite{RingNet, deng2019accurate, guo2020towards, wang20243d, DECA, danvevcek2022emoca, zielonka22mica, bai2021riggable}. While robust and semantically controllable, these approaches are inherently limited by the expressiveness of the underlying linear parametric model, often struggling to capture fine-grained details that lie beyond the model's subspace. HRN~\cite{lei2023hierarchical} employs an iterative refinement strategy, but it is prone to overfitting, which degrades the realism of the results. HiFace~\cite{chai2023hiface} attempts to address this by introducing a high-fidelity 3DMM with an expanded basis, yet it still cannot fully represent intricate facial features such as wrinkles and pores. 

Monocular 3D face geometry reconstruction remains a fundamental challenge~\cite{wang2021review}. The dominant paradigm relies on parametric representations, utilizing 3DMMs~\cite{blanz2023morphable, hifi3dface2021tencentailab, wang2022faceverse, FaceKit} or blendshapes~\cite{FLAME} to regress coefficients via deep learning~\cite{RingNet, deng2019accurate, guo2020towards, wang20243d, DECA, danvevcek2022emoca, zielonka22mica, bai2021riggable}. While semantically controllable, these methods are inherently constrained by the linear subspace, failing to capture high-frequency details. To address this, HRN~\cite{lei2023hierarchical} proposes a hierarchical iterative refinement strategy but suffers from optimization instability (e.g., overfitting artifacts). Similarly, HiFace~\cite{chai2023hiface} attempts to enhance expressiveness by expanding the high-fidelity 3DMM basis, yet it still struggles to resolve intricate micro-structures like wrinkles and pores.
% The second category, nonlinear facial reconstruction models, endeavors to surmount the inherent limitations of linear models in representing high-frequency geometric details, thereby enhancing both the accuracy and realism of reconstructions. Instead of relying on fixed linear bases, these approaches employ Deep Neural Networks (DNNs) or Graph Neural Networks (GNNs) to learn complex non-linear mappings directly from data, enabling the capture of subtle facial variations and personalized characteristics~\cite{tran2018nonlinear, tran2019towards,aliari2023face,ranjan2018generating,dib2024mosar}. Furthermore, recent advancements have explored Implicit Neural Representations (INRs) to model continuous facial geometry, offering superior resolution and topology flexibility~\cite{deluigi2023inr2vec, zheng2022imface, ImFace++}. By learning these non-linear transformations from 2D inputs to 3D geometry, such methods can recover richer, intricate details like wrinkles, leading to higher fidelity reconstructions. However, they typically demand extensive, high-quality 3D training data, and the resulting geometry often lacks the explicit semantic disentanglement offered by parametric models.
% Targeting high-fidelity texture synthesis, we enhance a linear parametric model~\cite{hifi3dface2021tencentailab} with DINOv3~\cite{simeoni2025dinov3} semantic priors. This yields a geometric scaffold sufficiently accurate for extracting visible texture cues, effectively supporting our generative pipeline without the complexity of non-linear model.
The second category, non-linear reconstruction models, endeavors to surmount the inherent limitations of linear models in representing high-frequency geometric details. Instead of relying on fixed linear bases, these approaches employ DNNs, GNNs~\cite{tran2018nonlinear, tran2019towards,aliari2023face,ranjan2018generating,dib2024mosar}, or Implicit Neural Representations (INRs)~\cite{deluigi2023inr2vec, zheng2022imface, ImFace++} to learn complex mappings directly from data. This enables the capture of subtle facial variations and intricate details like wrinkles, leading to higher fidelity reconstructions. However, they typically demand extensive, high-quality 3D training data, and the resulting geometry often lacks the explicit semantic disentanglement offered by parametric models. Targeting high-fidelity texture synthesis, we instead enhance a linear parametric model~\cite{hifi3dface2021tencentailab} with DINOv3~\cite{simeoni2025dinov3} semantic priors. This yields a geometric scaffold sufficiently accurate for extracting visible texture cues, effectively supporting our generative pipeline without the complexity of non-linear models.

\subsection{Facial texture reconstruction}

Monocular facial texture reconstruction has evolved from 3DMM-based parameter regression~\cite{deng2019accurate,smith2020morphable} to high-fidelity generative approaches.
The latter typically formulates texture generation as a UV-space completion problem. Early approaches utilized the editing and generative capabilities of Generative Adversarial Networks (GANs)~\cite{pix2pix2017,karras2019style, karras2020analyzing} to hallucinate invisible regions from multi-view inputs~\cite{deng2018uv,gecer2019ganfit,gecer2021ostec}. More recently, latent diffusion models~\cite{rombach2022high} have been introduced for this task; for instance, UV-IDM~\cite{li2024uv} and FreeUV~\cite{yang2025freeuv} employ diffusion processes in latent space to generate high-quality texture details. But these completion-based methods inextricably bake the illumination into the texture, rendering the reconstructed assets unsuitable for relighting or downstream applications.

To achieve disentangled, physically-based reconstruction, recent research seeks to replace linear texture bases with high-quality non-linear bases generated by GANs or Diffusion models. These methods typically rely on large-scale synthetic datasets~\cite{bai2023ffhq,albedogan,dai2025high} or high-quality scans from Light Stage devices~\cite{Lattas2022AvatarMe++,lattas2020avatarme,Lattas2023fitme,fitdiff,Papantoniou2023Relightify,dib2024mosar}, employing differentiable rendering for optimization.
While MoSAR~\cite{dib2024mosar} achieved complete texture reconstruction, its GAN-based backbone often suffers from mode collapse and over-smoothed results. Moreover, approaches relying on standard 2D differentiable rasterization often fail to explicitly handle occlusions, inadvertently baking artifacts from hair or glasses into the facial texture.
NextFace~\cite{nextface} utilizes differentiable ray tracing to explicitly model self-occlusion for high-quality material disentanglement; however, compared to rasterization-based methods, it incurs prohibitive computational overhead and lacks generalizability. 
UltraAvatar~\cite{zhou2024ultravatar} attempts to fine-tune early versions of diffusion models on small datasets, yet the generation quality remains limited, and the model loses the capability to generalize to in-the-wild images.

Fundamentally, these methods are constrained by a data-efficiency dilemma: synthetic data often lacks realism, while real Light Stage data is expensive and difficult to scale. In contrast, we utilize a modern rendering engine~\footnote{https://www.blender.org/} to augment limited real texture data. By leveraging the powerful generative priors of large-scale diffusion models~\cite{LongCat-Image, team2025zimage, labs2025flux1kontextflowmatching,wu2025qwenimagetechnicalreport}, we achieve high-fidelity generation of comprehensive material components with fewer than 100 real samples, seamlessly generalizing to in-the-wild images.
\section{Methods}
\label{sec:methods}

\begin{figure*}[t]
  \centering
  \includegraphics[width=0.95\linewidth]{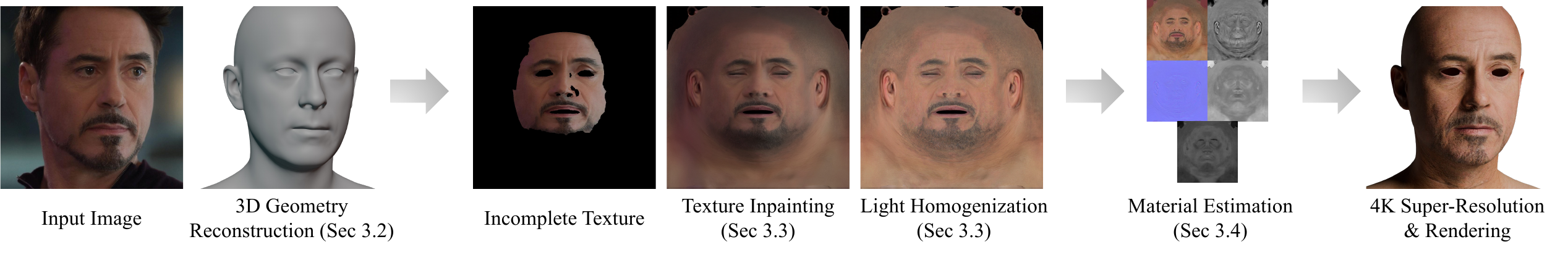}
  \vspace{-5pt}
  \caption{\textbf{Overview of our high-fidelity 3D avatar reconstruction pipeline.}
From a single image, we recover 3D geometry and an incomplete texture map, then repair and homogenize it for illumination. We subsequently predict PBR maps (albedo, roughness, specular, normal) and displacement, enabling photorealistic 4K rendering.}
  \label{fig:pipeline}
  \vspace{-5pt}
\end{figure*}

An overview of our method is illustrated in Fig.~\ref{fig:pipeline}.
Given a single in-the-wild input image, we first estimate the 3D face geometry and camera parameters, which are utilized to project the image into UV space to obtain an incomplete texture map (Sec.~\ref{sec:geometry}).
To reconstruct a complete and canonicalized texture representation, we introduce a shared pre-trained diffusion backbone as the core component.
Adapted via cascaded LoRA~\cite{hu2022lora} modules, this backbone is flexibly applied to a sequence of sub-tasks:
first, a texture inpainting stage completes missing regions caused by occlusion;
this is followed by a light homogenization step that removes baked-in illumination to establish a unified lighting baseline (Sec.~\ref{sec:inpaint_lighthom}).
Building on this, a multi-branch material estimation module utilizes the normalized texture to predict key intrinsic material parameters, including albedo, normal, roughness, specular, and displacement maps (Sec.~\ref{sec:material}).
Finally, all predicted maps are upscaled to 4K resolution by a super-resolution network, ensuring high-fidelity rendering output (Sec.~\ref{sec:sr}).

% --- START OF SECTION 3.1 ---

\subsection{Dataset Preparation}
\label{sec:dataset_prep}

\begin{figure}[t]
  \centering
  \begin{minipage}[t]{0.48\linewidth}
    \centering
    \includegraphics[width=\linewidth]{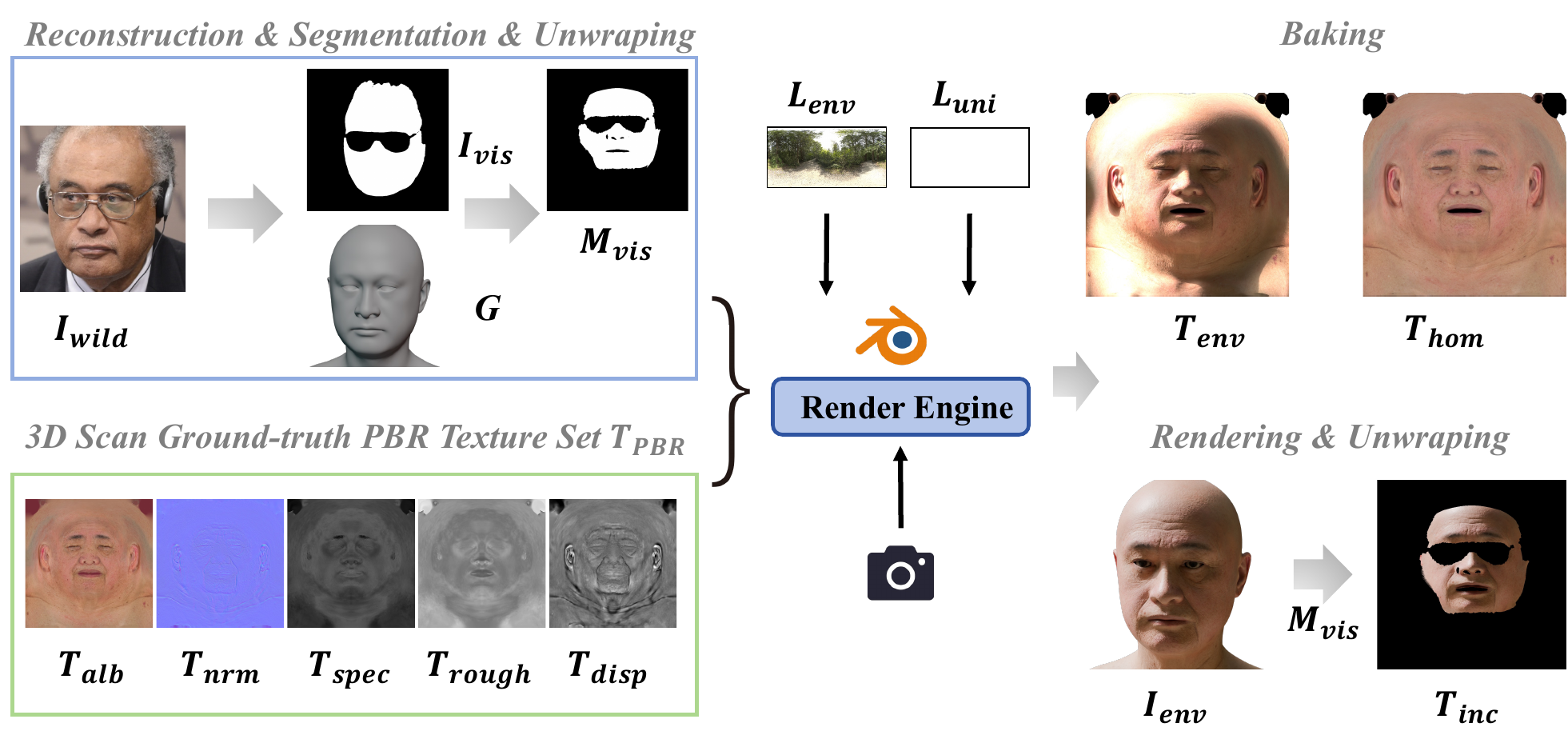}
    \caption{\textbf{Synthetic Data Generation Pipeline.} Starting from an image $I_{wild}$, we reconstruct geometry $\mathbf{G}$ and compute visibility mask $M_{vis}$. We assign high-quality PBR textures $T_{PBR}$ to $\mathbf{G}$.}
    \label{fig:data_pipeline}
  \end{minipage}
  \hfill
  \begin{minipage}[t]{0.48\linewidth}
    \centering
    \includegraphics[width=\linewidth]{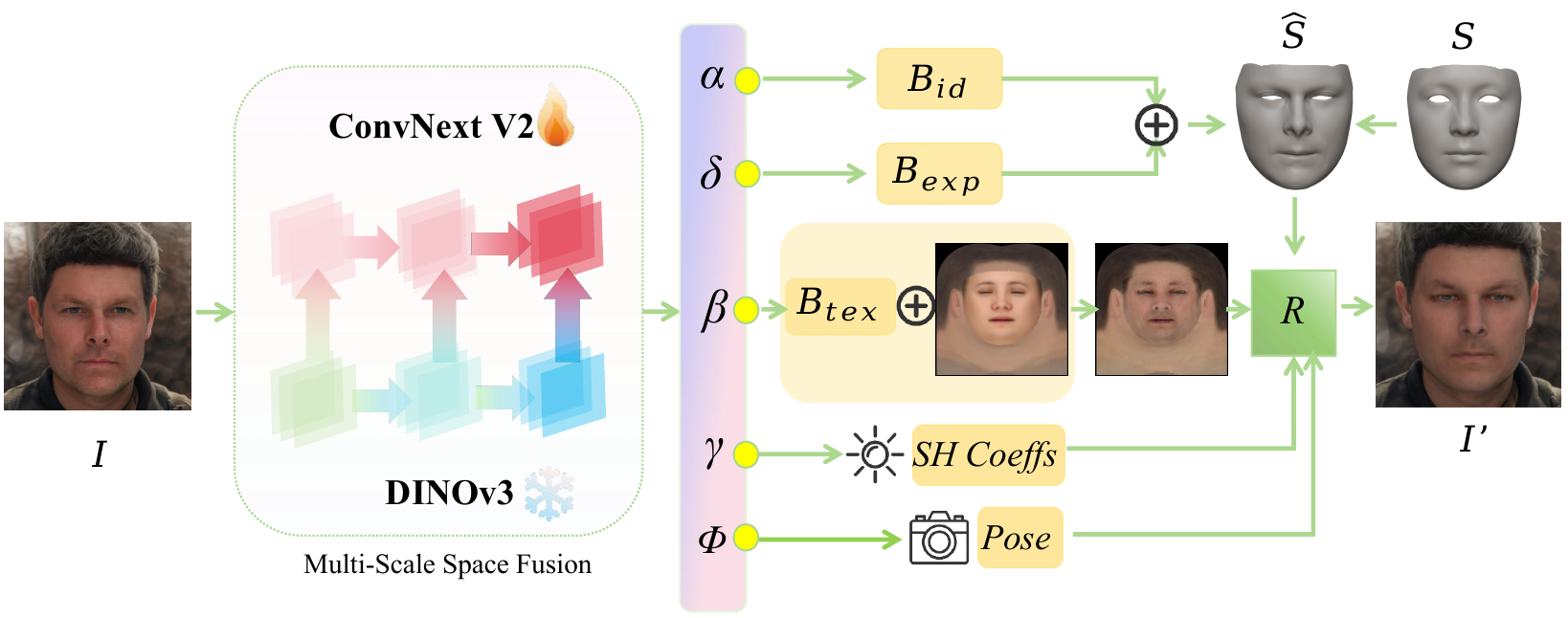}
    \caption{\textbf{Geometry estimation model.} Combining ConvNeXt V2 and DINOv3, our encoder integrates local details with semantic priors for geometry estimation.}
    \label{fig:geometry_pipeline}
  \end{minipage}
\end{figure}

Our training data are constructed by combining large-scale in-the-wild images
with a small set of high-quality scanned PBR textures, following the pipeline
illustrated in Fig.~\ref{fig:data_pipeline}.
We use FFHQ~\cite{karras2019style} and CelebAMask-HQ~\cite{CelebAMask-HQ}
as sources of in-the-wild facial images $I_{wild}$,
which provide diverse geometry, pose, expression, illumination, and occlusion.
These images are used to train the geometry reconstruction module
and to derive realistic visibility masks $M_{vis}$.
For material learning, we employ a compact scan dataset $D_{scan}$~\footnote{https://www.3dscanstore.com.}
containing fewer than 100 professional 3D face scans.
Each scan provides a complete set of physically-based material maps,
including albedo ($T_{alb}$), normal ($T_{nrm}$), roughness ($T_{rough}$),
specular ($T_{spec}$), and displacement ($T_{disp}$),
which together form the PBR texture set $T_{PBR}$.
Rather than relying on appearance diversity,
we treat these scans as physically accurate anchors.
As shown in Fig.~\ref{fig:data_pipeline},
we first reconstruct facial geometry $\mathbf{G}$ from $I_{wild}$
and compute the visibility mask $M_{vis}$ in UV space.
The scanned PBR textures $T_{PBR}$ are then assigned to $\mathbf{G}$
and rendered under diverse lighting $L_{env}$ to produce shaded textures $T_{env}$,
as well as under uniform lighting $L_{uni}$ to obtain light-homogenized targets $T_{hom}$.
To generate realistic incomplete inputs,
we render images $I_{env}$, re-unwrap them into UV space,
and multiply by $M_{vis}$ to obtain $T_{inc}$.
These paired data supervise texture inpainting, light homogenization,
and intrinsic material estimation.

% --- END OF SECTION 3.1 ---

% --- START OF SECTION 3.2 ---

\subsection{3D Geometry Reconstruction}
\label{sec:geometry}

Accurate 3D geometry forms the foundation of our high-fidelity avatar pipeline.
Following standard paradigms~\cite{deng2019accurate},
we regress the parameters of a 3D Morphable Model (3DMM) from a single image
using deep convolutional networks.
To enhance geometric fidelity beyond conventional linear models,
we introduce a multi-scale encoder architecture together with a robust hybrid
landmark supervision strategy.

We adopt the Hifi3D++ basis as our parametric face model.
The parameter vector
$\chi = \{\boldsymbol{\alpha}, \boldsymbol{\delta}, \boldsymbol{\beta},
\boldsymbol{\gamma}, \boldsymbol{\phi}\}$
represents identity
$\boldsymbol{\alpha} \in \mathbb{R}^{532}$,
expression
$\boldsymbol{\delta} \in \mathbb{R}^{45}$,
texture
$\boldsymbol{\beta} \in \mathbb{R}^{439}$,
Spherical Harmonic lighting
$\boldsymbol{\gamma} \in \mathbb{R}^{9}$,
and perspective camera pose $\boldsymbol{\phi}$.
The 3D shape $S$ and texture $T$ are modeled as:
\begin{equation}
\label{eq:3dmm}
\begin{aligned}
S(\boldsymbol{\alpha}, \boldsymbol{\delta}) &= \bar{S}
+ \mathbf{B}_{id}\boldsymbol{\alpha}
+ \mathbf{B}_{exp}\boldsymbol{\delta}, \\
T(\boldsymbol{\beta}) &= \bar{T}
+ \mathbf{B}_{tex}\boldsymbol{\beta},
\end{aligned}
\end{equation}
where $\bar{S}$ and $\bar{T}$ denote the mean shape and texture,
and $\mathbf{B}$ are the corresponding PCA bases.
The reconstructed image is obtained through a differentiable renderer
$\mathcal{R}$ as $I' = \mathcal{R}(\chi)$.

To capture both fine-scale geometric details and high-level semantic structure,
we propose a \textbf{Multi-Scale Space Fusion (MSSF) encoder}
(Fig.~\ref{fig:geometry_pipeline}).
MSSF combines a trainable ConvNeXt V2 backbone~\cite{woo2023convnext}
for local feature extraction
with a frozen DINOv3 model~\cite{simeoni2025dinov3}
that provides strong self-supervised semantic priors.
Multi-scale features from both branches are fused before parameter regression,
significantly improving reconstruction accuracy.
Despite using a linear decoder, this design enables our method to outperform
nonlinear approaches such as MoSAR~\cite{dib2024mosar}
in recovering subtle facial structures.

The network is trained using a self-supervised objective:
\[
\mathcal{L}
=
\lambda_{pho}\mathcal{L}_{pho}
+
\lambda_{lan}\mathcal{L}_{lan}
+
\lambda_{per}\mathcal{L}_{per}
+
\lambda_{reg}\mathcal{L}_{reg},
\]
where $\mathcal{L}_{pho}$ is a photometric loss weighted by a skin-attention mask
to handle occlusions,
$\mathcal{L}_{per}$ is a perceptual loss to improve visual fidelity,
and $\mathcal{L}_{reg}$ regularizes the 3DMM coefficients.
To address instability of standard 68-point landmark detectors around the mouth,
we introduce an \textbf{Enhanced Landmark Loss} $\mathcal{L}_{lan}$
based on an 88-point hybrid configuration,
which combines stable facial contour landmarks~\cite{Bulat_2017_ICCV}
with robust mouth keypoints from MediaPipe~\cite{lugaresi2019mediapipe}.

% --- END OF SECTION 3.2 ---
\begin{figure*}[t]
  \centering
  \includegraphics[width=1\linewidth]{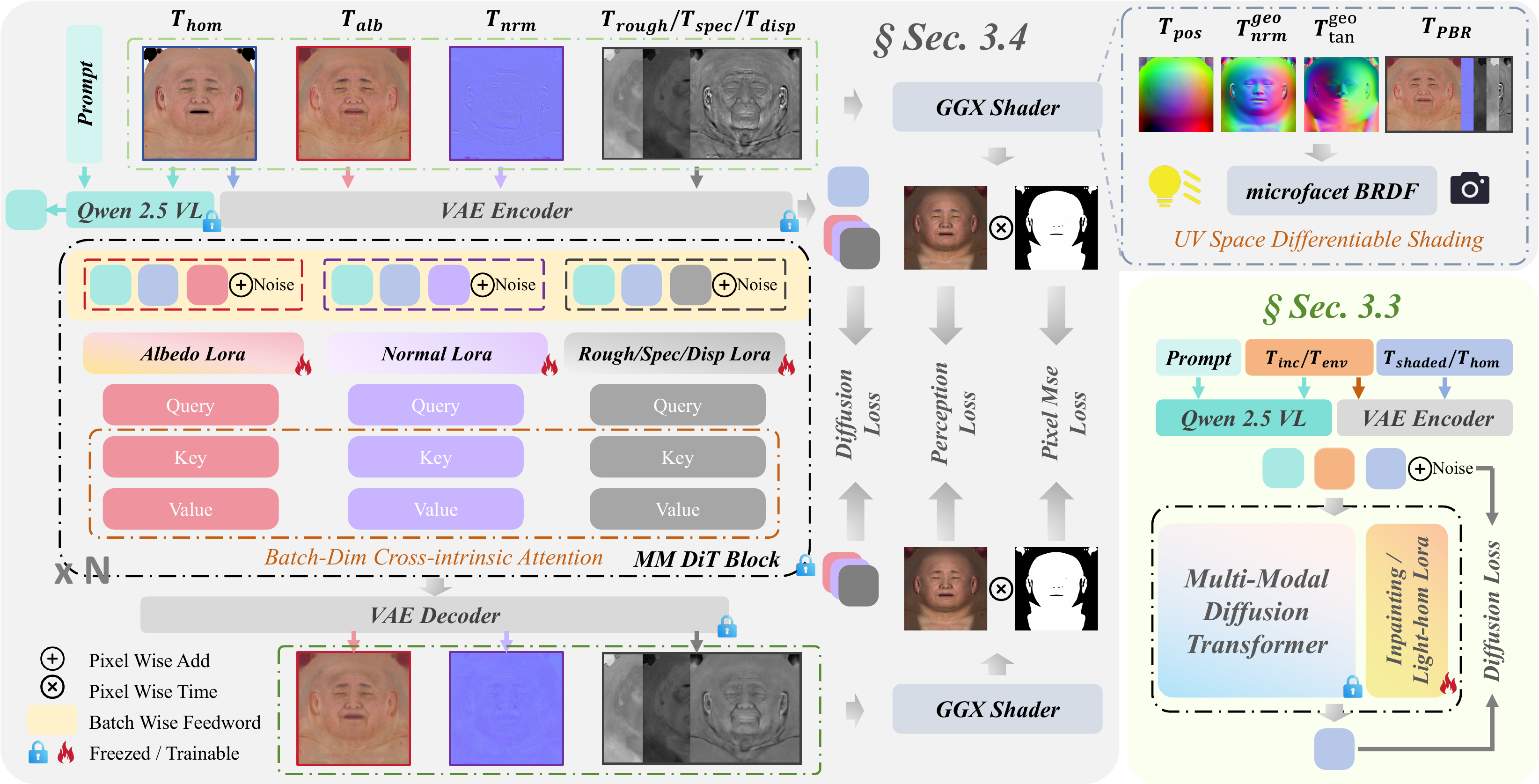}
\caption{
\textbf{Training pipeline for texture generation.}
A unified diffusion transformer uses task-specific LoRA modules for UV-space texture inpainting, light homogenization, and PBR material estimation.
Inpainting and homogenization (Sec.~\ref{sec:inpaint_lighthom}) use paired supervision.
For material estimation (Sec.~\ref{sec:material}), the homogenized texture $T_{\mathrm{hom}}$ feeds three parallel LoRA branches (albedo, normal, packed) interacting via cross-intrinsic attention to generate PBR maps, supervised by a differentiable GGX BRDF shader.}
  \label{fig:train_pipeline}
\end{figure*}

\vspace{-5pt}

\subsection{Texture Inpainting and Light Homogenization}
\label{sec:inpaint_lighthom}

Leveraging the paired data constructed in Sec.~\ref{sec:dataset_prep},
we perform texture completion and light homogenization using a unified
diffusion backbone operating in UV space, adapted via task-specific LoRA modules.
Both tasks are formulated as conditional diffusion processes that share
the same pre-trained transformer and differ only in their conditioning inputs
and supervision targets (Fig.~\ref{fig:train_pipeline}, right).
During training, the backbone is kept frozen and only the LoRA parameters are optimized.

\paragraph{Texture Inpainting.}
Texture inpainting aims to recover a complete UV texture from an incomplete input
$T_{\mathrm{inc}}$.
We formulate this task as conditional diffusion in latent space using the flow-matching objective.
Given a noise level $\sigma \in (0,1)$, the noisy latent is constructed as
$z_{\sigma} = (1-\sigma)\, z_0 + \sigma\, \epsilon$,
where $z_0$ denotes the latent encoding of the ground-truth texture and
$\epsilon \sim \mathcal{N}(0, \mathbf{I})$.
The model predicts the target velocity field $(\epsilon - z_0)$
by minimizing the flow-matching loss:
\begin{equation}
\label{eq:flow_loss}
\mathcal{L}_{\mathrm{FM}}
=
\mathbb{E}_{\sigma,\,\epsilon}
\left[
\left\|
f_{\mathrm{inp}}(z_{\sigma}, T_{\mathrm{inc}}, \sigma)
-
(\epsilon - z_0)
\right\|_2^2
\right].
\end{equation}
The incomplete texture $T_{\mathrm{inc}}$ is provided as conditioning,
enabling the model to infer missing regions while preserving visible content.

\paragraph{Light Homogenization.}
Although texture completion yields visually complete UV maps, illumination remains scene-dependent. To enable intrinsic material estimation, we introduce a light homogenization stage that maps shaded textures into a canonical lighting domain. Using the same diffusion backbone and flow-matching formulation, the conditioning input and supervision target differ: the model is conditioned on the shaded texture \(T_{\mathrm{env}}\) and supervised by its uniformly illuminated counterpart \(T_{\mathrm{hom}}\). Normalizing textures into a consistent lighting space simplifies and stabilizes subsequent material estimation.

\subsection{Physically-based Material Estimation}
\label{sec:material}

This stage estimates physically-based material properties from the light-homogenized texture $T_{\mathrm{hom}}$.
Our goal is to recover a complete PBR representation in UV space,
including albedo $T_{\mathrm{alb}}$, normal $T_{\mathrm{nrm}}$,
and a compact reflectance map
$T_{\mathrm{rsd}}$ that packs roughness $T_{\mathrm{rough}}$,
specular $T_{\mathrm{spec}}$, and displacement $T_{\mathrm{disp}}$.
A naïve approach is to train independent diffusion adapters for each material attribute.
While the semantic priors from the pre-trained backbone are effectively leveraged, independently generated material maps often lack physical consistency. Specifically, independently trained branches tend to overfit local high-frequency pixel variations, misinterpreting appearance cues as geometric structures, which leads to fragmented artifacts and spatially incoherent micro-structures.

\paragraph{Joint Diffusion with Cross-Intrinsic Attention.}
To enforce physical coherence across material modalities,
we adopt a joint diffusion strategy based on \emph{cross-intrinsic attention}.
We introduce three task-specific LoRA adapters,
$\Delta\theta_{\mathrm{alb}}$, $\Delta\theta_{\mathrm{nrm}}$, and
$\Delta\theta_{\mathrm{rsd}}$,
corresponding to albedo, normal, and reflectance-related properties,
while sharing a common diffusion backbone.
During training, the three material modalities are processed jointly by
stacking them along the batch dimension.
Within each transformer block, standard self-attention is replaced by a
cross-intrinsic attention mechanism that enables explicit information
exchange across modalities.
Queries are computed independently within each modality branch,
while keys and values are formed by concatenating features from all
material branches.
Let
$
\mathbf{K} = [\mathbf{k}_{\mathrm{alb}},\mathbf{k}_{\mathrm{nrm}},\mathbf{k}_{\mathrm{rsd}}],
\quad
\mathbf{V} = [\mathbf{v}_{\mathrm{alb}},\mathbf{v}_{\mathrm{nrm}},\mathbf{v}_{\mathrm{rsd}}]
$
denote the concatenated keys and values across modalities.
Given query features $\mathbf{q}_i$ from modality $i$,
cross-intrinsic attention outputs the updated token features
$\mathbf{h}_i$ as
\begin{equation}
\mathbf{h}_i = \mathrm{Attn}(\mathbf{q}_i, \mathbf{K}, \mathbf{V}),
\mathrm{Attn}(\mathbf{q},\mathbf{K},\mathbf{V})
=
\mathrm{softmax}\!\left(\frac{\mathbf{q}\mathbf{K}^{\mathsf{T}}}{\sqrt{d}}\right)\mathbf{V}.
\label{eq:cross_intrinsic_attn}
\end{equation}
This design allows each material branch to directly attend to
complementary geometric and reflectance cues from the others,
encouraging spatially aligned fine-scale structures while preserving
domain-specific adaptation through separate LoRA parameters.

\paragraph{UV-Space Differentiable Shader.}
Joint diffusion enforces spatial alignment but does not
guarantee physically meaningful decomposition.
We therefore introduce a UV-space differentiable BRDF shading loss that
explicitly couples material prediction with image formation.
During training, the clean latent estimate
$\hat{\mathbf{z}}_{0}$ is reconstructed from the noisy state
$\mathbf{z}_{\sigma}$ and the predicted velocity field
$\hat{\mathbf{v}}_{\theta}$ (Eq.~\ref{eq:flow_loss}) as:
\begin{equation}
\label{eq:x0_recon}
\hat{\mathbf{z}}_{0}
=
\mathbf{z}_{\sigma}
-
\sigma\,\hat{\mathbf{v}}_{\theta}(\mathbf{z}_{\sigma},\sigma,\mathbf{c}).
\end{equation}
Decoding $\hat{\mathbf{z}}_{0}$ through the VAE yields the predicted material maps
$\hat{T}_{\mathrm{alb}}$, $\hat{T}_{\mathrm{nrm}}$, and $\hat{T}_{\mathrm{rsd}}$.
To ensure stable and physically grounded supervision,
we decouple texture decoding from subject-specific geometry.
A fixed template face is used, for which we precompute its world-space
position $\tilde{T}_{\mathrm{pos}}$, geometric normal
$\tilde{T}^{\mathrm{geo}}_{\mathrm{nrm}}$, and geometric tangent
$\tilde{T}^{\mathrm{geo}}_{\mathrm{tan}}$.
Given these geometric priors, shading is evaluated using a GGX microfacet
shader:
\begin{equation}
\label{eq:ggx_render}
\hat{T}_{\mathrm{shaded}}
=
\mathcal{S}_{\mathrm{GGX}}
\!\left(
\hat{T}_{\mathrm{alb}},
\hat{T}_{\mathrm{nrm}},
\hat{T}_{\mathrm{rough}},
\hat{T}_{\mathrm{spec}},
\hat{T}_{\mathrm{disp}};
\tilde{T}_{\mathrm{pos}},
\tilde{T}^{\mathrm{geo}}_{\mathrm{nrm}},
\tilde{T}^{\mathrm{geo}}_{\mathrm{tan}},
L, V
\right),
\end{equation}
where $L$ and $V$ denote sampled directional light and view directions.
Implementation details of $\mathcal{S}_{\mathrm{GGX}}$ are provided in the Appendix A.1.
During training, camera viewpoints are sampled using a hybrid strategy:
with probability $0.5$, views are sampled uniformly along a $360^\circ$
azimuth at a fixed radius with random vertical perturbation; otherwise,
a near-frontal view with small spatial jitter is used.
Directional lighting is initialized from the camera direction and randomly
perturbed, and a low-intensity uniform ambient term sampled from
$[0.15, 0.3]$ is added to prevent degenerate shadows.

\paragraph{Training Objective.}
The material estimation network is optimized using a hybrid objective that
combines diffusion supervision with rendering-based constraints:
\begin{equation}
\mathcal{L}
=
\mathcal{L}_{\mathrm{diff}}
+
\lambda_{\mathrm{img}}
\,
\|\hat{T}_{\mathrm{shaded}} - T_{\mathrm{shaded}}\|_2^2
+
\lambda_{\mathrm{lpips}}
\,
\mathrm{LPIPS}(\hat{T}_{\mathrm{shaded}}, T_{\mathrm{shaded}}).
\end{equation}
By coupling flow-matching diffusion with physically grounded re-rendering
losses, the model is guided to produce material maps that exhibit
consistent physical behavior under novel illumination.

\subsection{Super-Resolution}
\label{sec:sr}
As a final step, we apply super-resolution to obtain 4K UV texture maps.
We treat super-resolution as a post-processing module
and fine-tune a Real-ESRGAN~\cite{wang2021realesrgan}
model to upscale all predicted texture maps from 1K to 4K.
 \section{Experiments}
 \label{sec:exps}

\begin{figure}[t]
    \centering
    \begin{minipage}[c]{0.49\linewidth} 
        \centering
        \includegraphics[width=\linewidth]{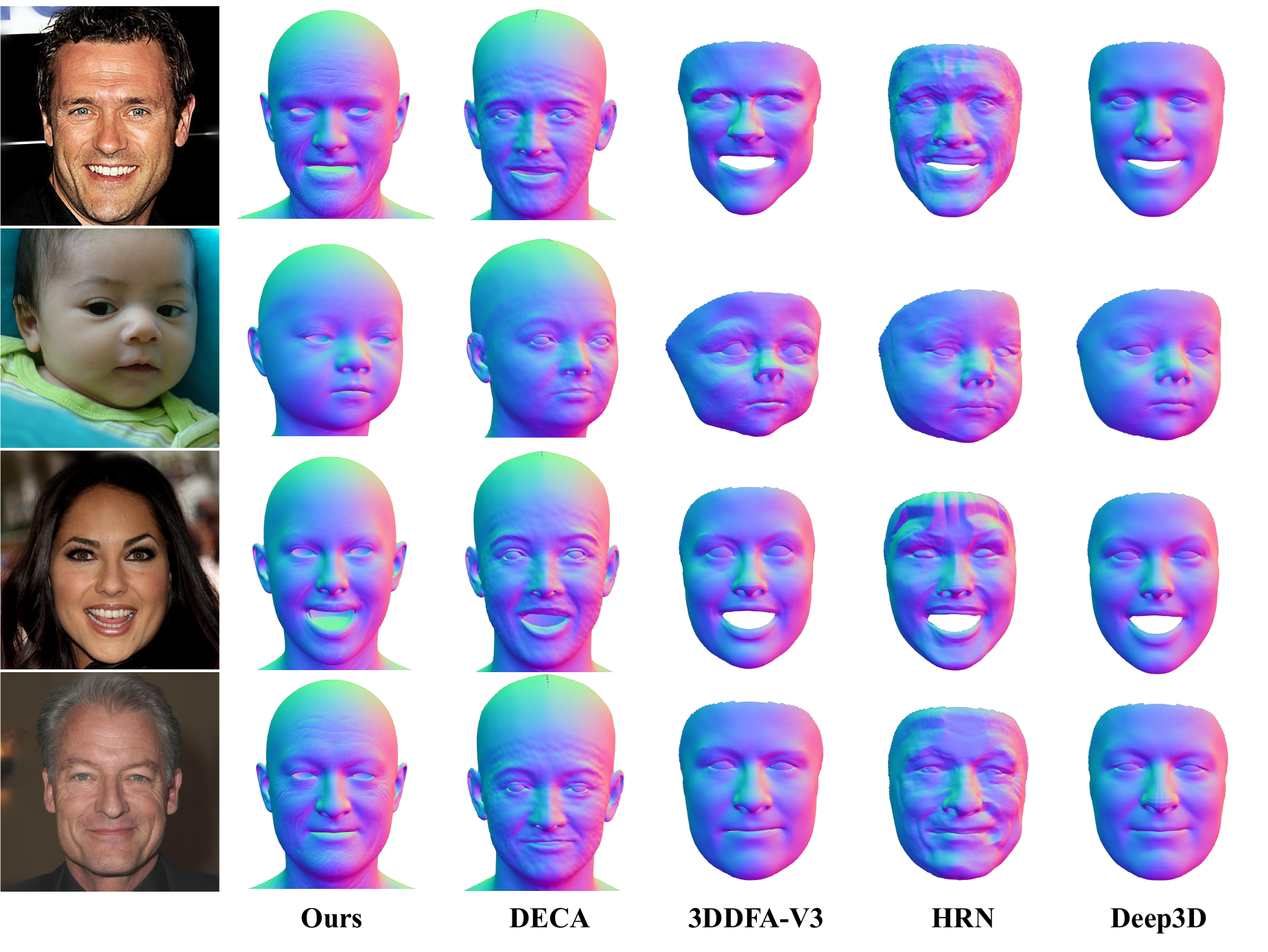}
        \vspace{-10pt}
        \caption{Comparison of 3D face reconstruction methods.}
        \label{fig:small_comparison}
    \end{minipage}
    \hfill % 左右之间的弹性空格
    \begin{minipage}[c]{0.49\linewidth}
        \centering
        \captionof{table}{Reconstruction errors (NMSE) on the REALY benchmark. Lower is better.}
        \label{tab:realy_benchmark}
        \resizebox{\linewidth}{!}{
            \begin{tabular}{l|c|c|c|c|c}
            \toprule
            \multirow{2}{*}{Method} & \multicolumn{4}{c|}{Error (mm)} & \multirow{2}{*}{All} \\
            \cline{2-5}
             & Nose & Mouth & Forehead & Cheeks & \\
            \midrule
            DECA & 1.697 $\pm$ 0.355 & 2.516 $\pm$ 0.839 & 2.394 $\pm$ 0.576 & 1.479 $\pm$ 0.535 & 2.010 \\
            Deep3D & 1.719 $\pm$ 0.354 & 1.368 $\pm$ 0.439 & 2.015 $\pm$ 0.449 & 1.528 $\pm$ 0.501 & 1.657 \\
            HRN & 1.722 $\pm$ 0.330 & 1.357 $\pm$ 0.523 & 1.995 $\pm$ 0.476 & 1.072 $\pm$ 0.333 & 1.537 \\
            HiFace (w/o syn) & 1.227 $\pm$ 0.407 & 1.787 $\pm$ 0.439 & 1.454 $\pm$ 0.382 & 1.762 $\pm$ 0.436 & 1.558 \\
            MoSAR & 1.499 $\pm$ 0.366 & 1.424 $\pm$ 0.462 & 1.950 $\pm$ 0.559 & 1.128 $\pm$ 0.303 & 1.500 \\
            3DDFA-V3 & 1.584 $\pm$ 0.308 & 1.237 $\pm$ 0.375 & 1.809 $\pm$ 0.394 & 1.110 $\pm$ 0.328 & 1.435 \\
            HiFace & 1.036 $\pm$ 0.280 & 1.450 $\pm$ 0.413 & 1.324 $\pm$ 0.334 & 1.291 $\pm$ 0.362 & 1.275 \\
            \midrule
            \textbf{Ours} & 1.619 $\pm$ 0.381 & 1.376 $\pm$ 0.477 & 1.784 $\pm$ 0.464 & 1.181 $\pm$ 0.402 & 1.490 \\
            w/o Enhanced $\mathcal{L}_{lan}$ & 1.457 $\pm$ 0.341 & 1.670 $\pm$ 0.529 & 1.803 $\pm$ 0.453 & 1.140 $\pm$ 0.405 & 1.518 \\
            w/o MSSF & 1.728 $\pm$ 0.441 & 1.785 $\pm$ 0.514 & 2.179 $\pm$ 0.610 & 1.530 $\pm$ 0.548 & 1.806 \\
            Baseline & 2.592 $\pm$ 0.510 & 2.664 $\pm$ 0.528 & 2.273 $\pm$ 0.563 & 2.524 $\pm$ 0.474 & 2.514 \\ 
            \bottomrule
            \end{tabular}
        }
    \end{minipage}
\end{figure}

\textbf{Implementation Details.} For geometry reconstruction, we train a ConvNeXt V2 Base backbone augmented with a frozen DINOv3 encoder for 50 epochs using a batch size of 128.
For texture generation, we adopt LongCat-Image-Edit~\cite{team2025longcat} as the unified diffusion transformer (DiT) backbone.
LoRA adapters (rank 32) are injected into the attention and feedforward projection layers of the DiT. %(see Appendix~\ref{app:lora_layers} for details).
Each adapter contains approximately 91M parameters, accounting for only 0.7\% of the backbone. We use JoyCaption\footnote{\url{https://github.com/fpgaminer/joycaption}} to generate captions for ground-truth data at each training stage, while inference combines captions from the input image with fixed templates for
intermediate tasks.
All models are trained on 8 NVIDIA H100 80GB GPUs using BF16 precision and
the AdamW optimizer~\cite{adamw} with a learning rate of $1\times10^{-4}$.
The inpainting and light-homogenization LoRAs are trained for 20k steps with a batch size of 64.
The intrinsic material LoRAs follow a two-stage schedule, consisting of 10k steps of independent training followed by 10k steps of joint training, using a batch size of 8 and loss weights $\lambda_{img}=0.5$ and
$\lambda_{lpips}=0.1$.

\subsection{Geometry Reconstruction}
\label{exp:geo_recon}

\paragraph{Quantitative and qualitative analysis.}
We evaluate our geometry reconstruction on the REALY benchmark~\cite{chai2022realy},
which provides high-quality 3D scans and images of 100 subjects.
Following the standard protocol, we report the Normalized Mean Squared Error (NMSE)
between reconstructed and ground-truth meshes over four facial regions:
nose, mouth, forehead, and cheeks.
Quantitative results are shown in Tab.~\ref{tab:realy_benchmark}.
Our method achieves the third-best overall performance with an NMSE of 1.490.
Notably, despite relying on a linear 3DMM representation,
our approach outperforms MoSAR~\cite{dib2024mosar},
which adopts a more expressive nonlinear geometry model.

While our method trails slightly behind 3DDFA-V3~\cite{wang20243d} and HiFace~\cite{chai2023hiface} numerically,
it is important to note that HiFace heavily relies on large-scale synthetic images
and real 3D meshes for training.
Under comparable settings that restrict training to in-the-wild images,
HiFace exhibits inferior performance compared to our method.
Similarly, although 3DDFA-V3 achieves marginally lower errors using
segmentation-based landmark clustering,
qualitative inspection reveals a tendency to overfit the input image.
As shown in Fig.~\ref{fig:small_comparison},
3DDFA-V3 produces unstable geometric bumps on smooth regions
and fails to recover fine wrinkle details.
We further compare our method with DECA~\cite{DECA},
HRN~\cite{lei2023hierarchical}, and Deep3D~\cite{deng2019accurate}.
Benefiting from the estimated displacement and normal maps, our method faithfully reconstructs high frequency facial details,
such as forehead wrinkles, eye region creases, and smile lines.
In contrast, HRN suffers from severe overfitting artifacts,
Deep3D lacks fine-scale details,
and DECA often generates wrinkles that are misaligned with the input image.

\vspace{-5pt}

\paragraph{Ablation Study.} As shown in Tab.~\ref{tab:realy_benchmark} bottom, adding the enhanced $\mathcal{L}_{lan}$ to a ConvNeXt baseline improves mouth-region accuracy. Furthermore, our MSSF encoder achieves comprehensive improvements across all metrics, underscoring the critical role of DINOv3's semantic priors.

\subsection{Texture Reconstruction}

\begin{figure*}[t]
    \centering
    \includegraphics[width=0.65\linewidth]{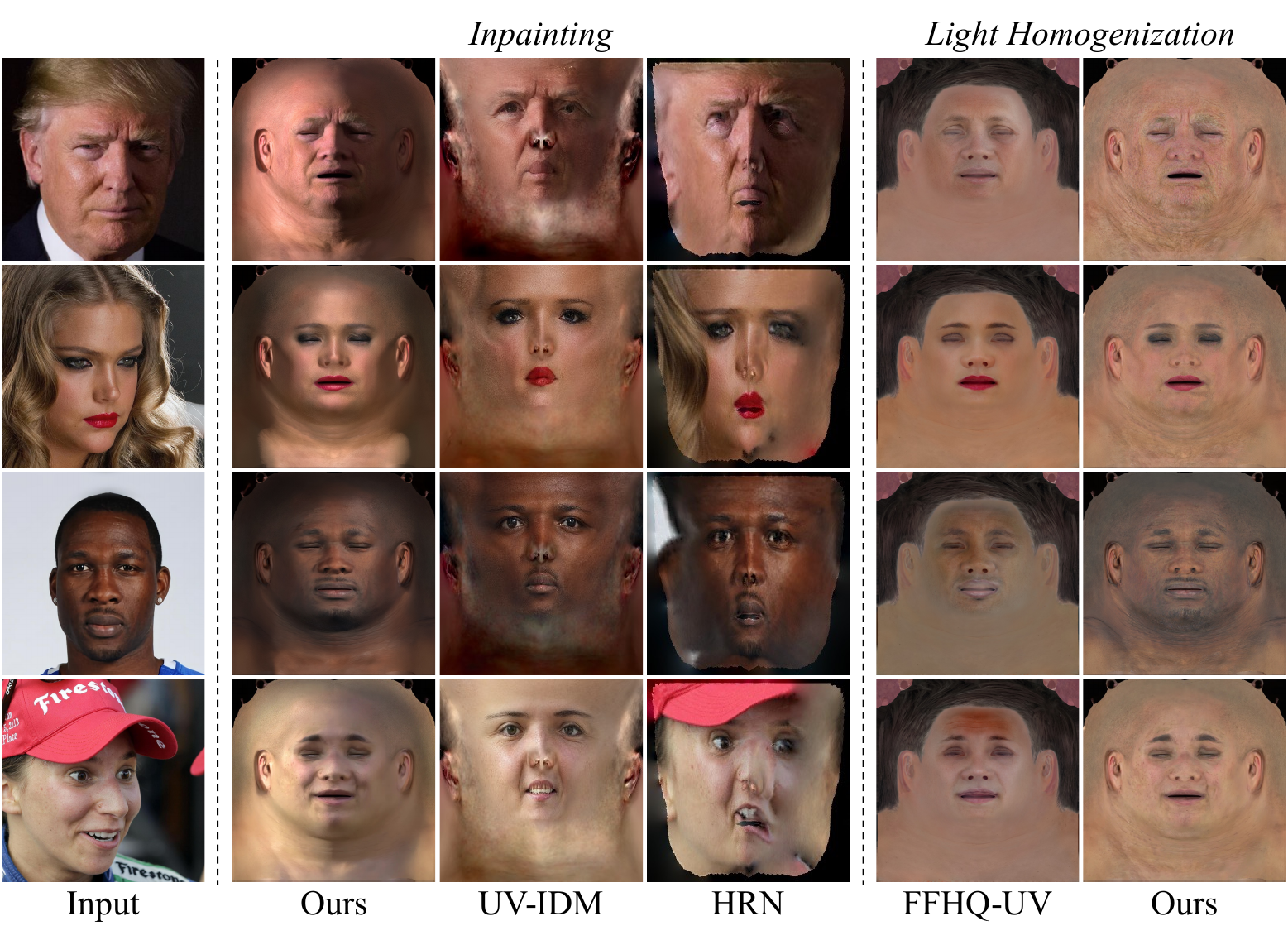}
    \vspace{-10pt}
\caption{\textbf{Texture inpainting and light homogenization.} Left: inputs. Middle: HRN shows baked-in occlusions while UV-IDM lacks fine detail; our method removes occlusions and restores high-frequency detail. Right: unlike FFHQ-UV, our homogenization yields uniformly lit, detail-rich textures for intrinsic decomposition.}
    \label{fig:tex_comparison}
\vspace{-3pt}
\captionof{table}{Quantitative evaluation of texture inpainting and light homogenization.}
\label{tab:texture_stage_metrics}
\scriptsize
\setlength{\tabcolsep}{3.2pt}
\renewcommand{\arraystretch}{0.9}
\begin{minipage}[t]{0.58\linewidth}
\centering
\textbf{(a) Texture Inpainting}\\[-1pt]
\resizebox{0.95\linewidth}{!}{%
\begin{tabular}{lcccc}
\toprule
Method & PSNR$\uparrow$ & SSIM$\uparrow$ & LPIPS$\downarrow$ & CSIM$\uparrow$ \\
\midrule
UV-IDM & 19.39 & 0.6149 & 0.1674 & 0.269 \\
HRN & 15.70 & 0.6193 & 0.2982 & 0.441 \\
Ours & \textbf{22.44} & \textbf{0.8003} & \textbf{0.0621} & \textbf{0.540} \\
\bottomrule
\end{tabular}
}
\end{minipage}
\hfill
\begin{minipage}[t]{0.36\linewidth}
\centering
\textbf{(b) Light Homogenization}\\[-1pt]
\resizebox{0.8\linewidth}{!}{%
\begin{tabular}{lcc}
\toprule
Method & CSIM$\uparrow$ & BS$\downarrow$ \\
\midrule
FFHQ-UV & 0.1340 & 5.738 \\
Ours & \textbf{0.4667} & \textbf{3.963} \\
\bottomrule
\end{tabular}
}
\end{minipage}
\vspace{-8pt}
\end{figure*}

\paragraph{Texture Inpainting and Light Homogenization}
Fig.~\ref{fig:tex_comparison} and Tab.~\ref{tab:texture_stage_metrics} demonstrate the effectiveness of our pipeline in handling occlusions and illumination.
For texture inpainting (middle block), we compare against UV-IDM~\cite{li2024uv} and HRN~\cite{lei2023hierarchical}.
Due to its reliance on an overfitting projection loss without explicit occlusion modeling, HRN tends to bake external objects into the UV texture.
As shown in the second and fourth rows, hair strands and hats are erroneously preserved, and large pose-induced self-occlusions are poorly handled.
UV-IDM avoids projection artifacts but struggles to preserve high-frequency identity details, exhibiting noticeable spotting artifacts (e.g., at the temples in the first row).
In contrast, our inpainting module produces clean and realistic skin textures in occluded regions while preserving identity-specific details.
Moreover, the resulting illumination appears more volumetric, benefiting from our integration of physically based rendering into the data generation process.

For light homogenization (right block), we compare against FFHQ-UV~\cite{bai2023ffhq}.
As shown in the third row, FFHQ-UV retains scene-dependent illumination, including specular highlights on the forehead and shadows around facial features, and is further affected by occlusions (e.g., the red artifact in the fourth row).
Our method effectively suppresses such illumination and occlusion artifacts, producing flat, uniformly lit textures that preserve fine appearance cues, including eyebrows, makeup, facial redness (rows 1, 2, and 4), as well as skin tone and beard structure (row 3).
This provides a robust and consistent input for subsequent PBR material estimation.

\paragraph{Material estimation.}
We compare our method with state-of-the-art 3D avatar reconstruction methods: MoSAR~\cite{dib2024mosar}, FitMe~\cite{Lattas2023fitme}, and Relightify~\cite{Papantoniou2023Relightify}. Since their official implementations are unavailable, we utilize test cases from~\cite{dib2024mosar} for a fair visual comparison.

\begin{figure*}[t]
    \centering
    \includegraphics[width=1\linewidth]{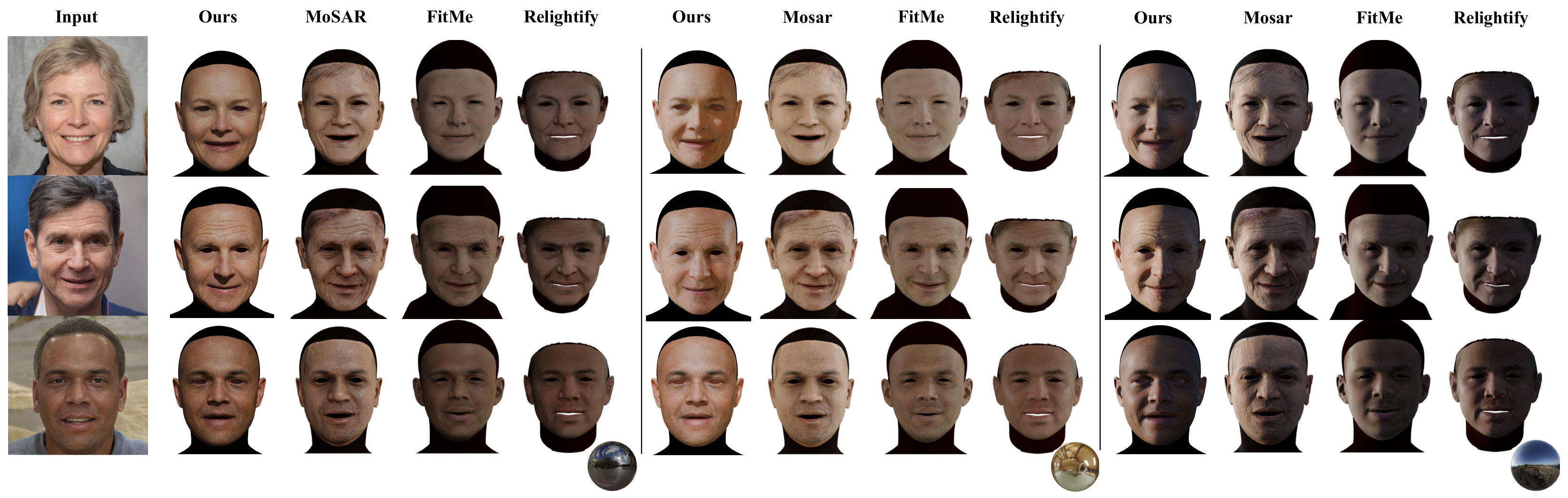} 
    \caption{\textbf{Qualitative comparison of avatar relighting under novel environments, against MoSAR, FitMe and Relightify.}
MoSAR yields waxy, desaturated skin and visible artifacts; FitMe and Relightify show poor albedo estimates and often bake hair into textures. Our method better disentangles lighting and occlusion, producing photorealistic, detailed avatars with accurate skin tones.}
    \label{fig:relighting}
\end{figure*}

\begin{figure*}[t]
  \centering
  \includegraphics[width=0.95\linewidth]{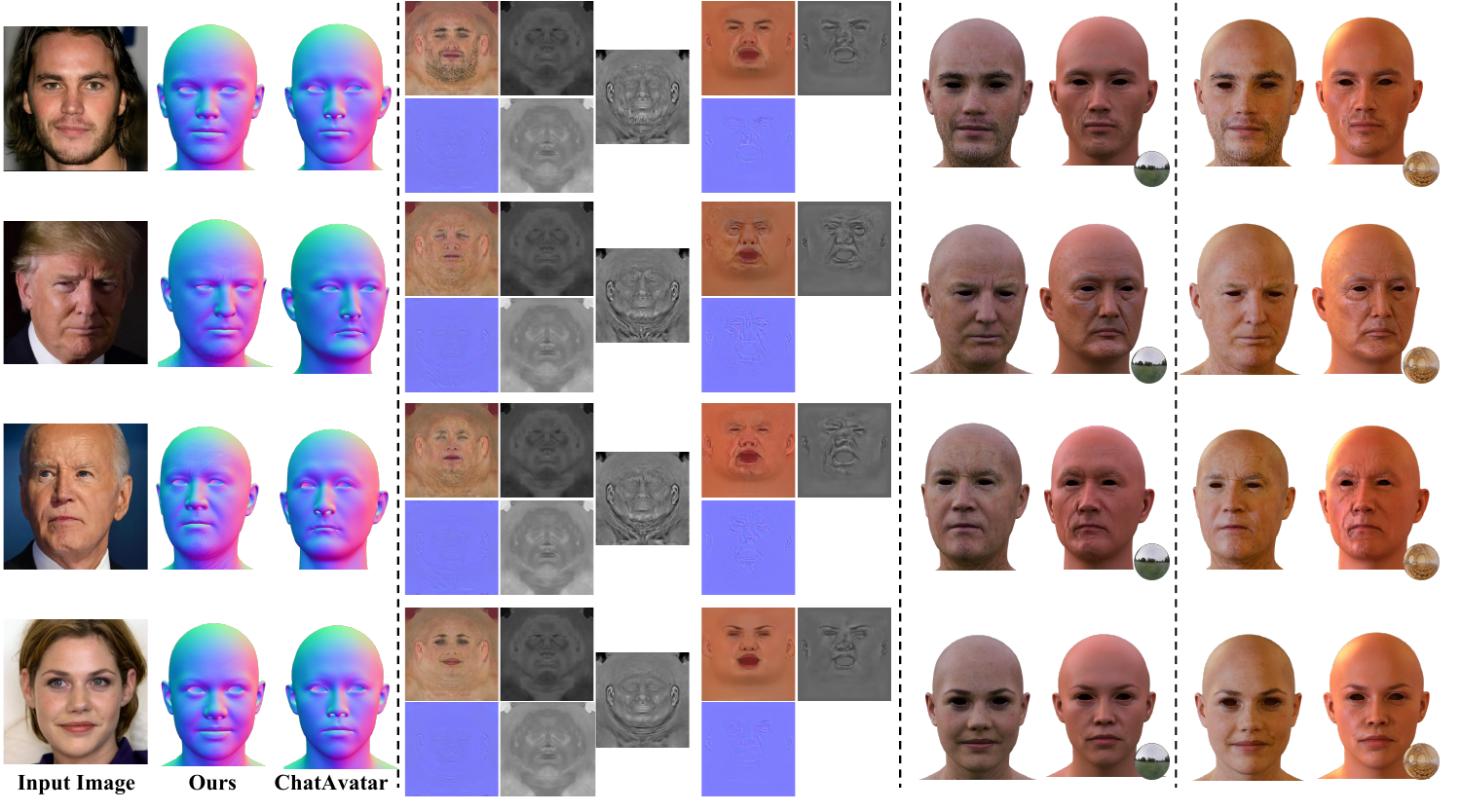} 
\caption{\textbf{Comparison with ChatAvatar.}
\textbf{Geometry (Left):} our method recovers high-frequency details (wrinkles, nasolabial folds) while the baseline over-smooths.
\textbf{Materials (Middle):} the baseline provides Albedo/Normal/Specular only; we predict full PBR including Roughness and Displacement.
\textbf{Relighting (Right):} under novel HDRIs, our relighting better preserves identity and photorealism.
}
  \label{fig:chatavatar}
  \vspace{-5pt}
\end{figure*}

As shown in Figures~\ref{fig:relighting} and~\ref{fig:chatavatar}, our pipeline demonstrates superiority in three key aspects.
First, regarding fidelity and skin tone (Fig.~\ref{fig:relighting}), MoSAR suffers from GAN-induced waxy over-smoothing and a systematic bias towards pale skin, failing to recover the dark pigmentation in Row 3. Similarly, FitMe and Relightify exhibit unnatural albedo estimation. In contrast, our method faithfully preserves authentic skin texture and intrinsic coloration.
Second, concerning occlusion robustness, rasterization-based methods like MoSAR struggle to separate skin from occluders, often baking hair into the texture (forehead in Rows 1-2) or generating artifacts (left temple in Row 3). Our UV-space physical constraints and generative priors effectively synthesize clean, artifact-free skin in occluded areas.
Finally, our approach even outperforms the closed-source commercial ChatAvatar, providing superior geometric details and material completeness—as seen in the beard in the first row, as well as the facial spots and wrinkle details in the second and third rows.

\paragraph{Ablation Study.}

\begin{figure*}[t]
  \centering
  \begin{subfigure}[b]{0.48\linewidth}
    \centering
    \includegraphics[width=\linewidth]{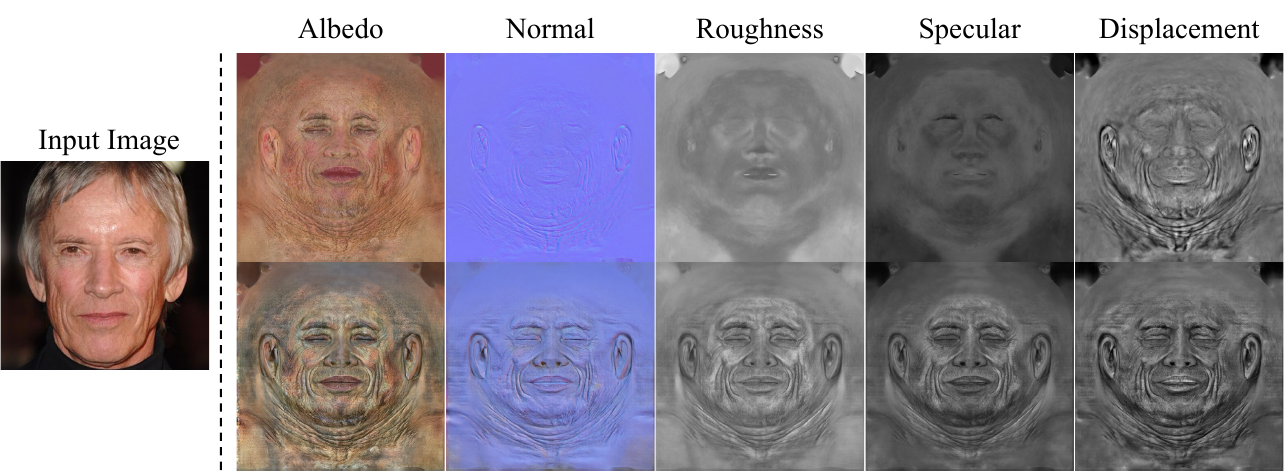}
\caption{\textbf{Ablation on light homogenization.}
\textbf{Top (Ours):} Homogenization ensures fast convergence to clean, physically disentangled materials.
\textbf{Bottom (w/o):} Without it, the solution space expands and the texture→material mapping becomes ill-posed; even after \textbf{40k steps}, the model fails to reach a physically plausible result. Shadows are mistaken for dark albedo (burnt artifacts) and geometric noise (noisy normals), reflecting sensitivity to lighting variance.
}

\label{fig:ablation_light}
  \end{subfigure}
  \hfill
  \begin{subfigure}[b]{0.48\linewidth}
    \centering
    \includegraphics[width=\linewidth]{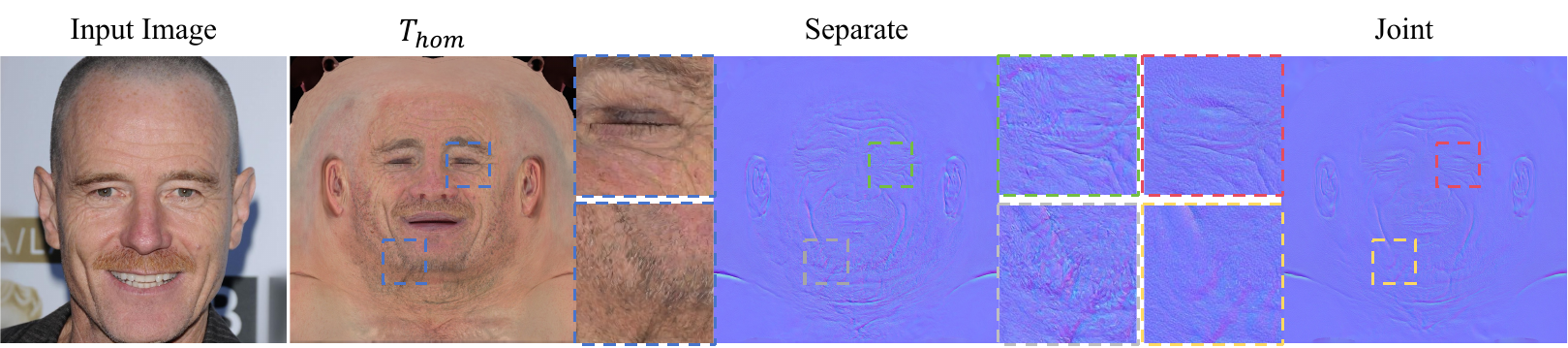}
\caption{\textbf{Visual ablation on material estimation.} Normals from independent LoRAs (Separate) vs our multi-branch (Joint). \textbf{Separate:} Lacking physical and cross-modal constraints, it overfits high-frequency detail—mistaking skin texture and stubble for geometry, causing fragmented crow’s feet and grainy cheeks. \textbf{Joint:} With Cross-Intrinsic Attention and a differentiable BRDF loss, we separate pigmentation from depth, producing coherent wrinkles and smooth, physically plausible skin while suppressing non-geometric noise.}
    \label{fig:ablation_normal}
  \end{subfigure}
  \caption{
    Texture Reconstruction ablation study.
  }
\end{figure*}

We first examine the necessity of the light homogenization stage in Fig.~\ref{fig:ablation_light}.
The top row shows results from the full pipeline, while the bottom row removes light homogenization and performs material estimation directly on the shaded inpainted texture. From a learning perspective, light homogenization acts as a crucial normalization step that significantly constrains the solution space.
Without it, the network faces a much harder one-to-many inverse rendering problem, requiring simultaneous material hallucination and implicit decomposition of unknown, complex lighting from a single texture map.
As shown in the bottom row, this expanded solution space leads to severe optimization instability: the model fails to recover valid material properties and instead converges to degenerate solutions, where high-contrast shading is baked into the albedo and misinterpreted as
high-frequency geometric noise in the normal and displacement maps. These results confirm that explicit lighting normalization is essential for stabilizing training and achieving physically disentangled reconstruction~\cite{li2025near}. 

We further analyze the qualitative impact of our multi-branch LoRA-based joint material estimation strategy in Fig.~\ref{fig:ablation_normal}.
A naive approach using separate LoRAs leads to modality-specific overfitting: as shown in the \emph{Separate} column, the network over-interprets high-frequency signals as geometric deformation, producing
severely noisy normal maps. In particular, stubble and skin pores are reconstructed as sharp geometric
spikes (gray box), while wrinkles appear as fragmented scratches rather than continuous folds (green box).
In contrast, our joint strategy combines Cross-Intrinsic Attention with a UV-space differentiable BRDF shading loss, enabling explicit inter-modal coupling and enforcing physical consistency.
This constraint suppresses geometry hallucination from surface pigmentation and restricts geometric variation to shading-consistent structures.
As shown in the \emph{Joint} column, this yields structurally coherent geometry, with continuous crow’s feet (red box) and smooth skin regions
(yellow box).

\section{Conclusions}
\label{sec:conclusions}

In this work, we present a novel framework for high-fidelity 3D avatar reconstruction from a single in-the-wild image. By synergizing modern rendering workflows with the robust priors of large-scale diffusion models, our approach effectively overcomes the data scarcity and generalization bottlenecks inherent in this task. Building upon a robust geometric alignment enhanced by DINOv3 priors, we introduce a unified diffusion-based pipeline for texture generation, facilitated by cascaded LoRA modules. This pipeline, through texture restoration, illumination homogenization, and intrinsic material estimation guided by a Cross-Intrinsic Attention mechanism combined with UV-space differentiable BRDF shading loss, successfully disentangles intrinsic materials from complex illumination without relying on screen-space supervision. Trained on fewer than 100 real 3D scans, our framework achieves state-of-the-art performance in recovering photorealistic, relightable, and geometrically consistent avatars.

% \paragraph{Limitation}
% Our method faces three limitations. First, recovering fine details behind semi-transparent occlusions (e.g., glasses in Fig.~\ref{fig:limitations}) remains challenging. Since the underlying texture is distorted or masked, the model tends to hallucinate generic, over-smoothed features rather than recovering identity-specific details (e.g., crow's feet). We plan to augment training with synthetic accessories to leverage generative priors for occlusion removal. Second, the multi-stage pipeline is computationally intensive, requiring $\sim$4 minutes on a single H100 GPU. We aim to address this via distillation strategies~\cite{liu2025decoupled,jiang2025distribution,team2025zimage}. Finally, the linear 3DMM basis and LoRA fine-tuning limit geometric expressiveness and open-domain editability. Future work will explore non-linear representations and regularization techniques to enhance fidelity and preserve model versatility.

\paragraph{Limitations.}
Our method targets relightable face and head PBR assets; it does not reconstruct full bodies, dynamic hair, clothing, or accessories. Treating hair, glasses, and hands as occlusions improves face-centric material recovery, but may lose identity-specific details behind occluders and fail under extreme expressions. Future work will explore complete heads, stronger nonlinear or 3DGS-based representations, larger open PBR datasets, and bias-aware evaluation.

\paragraph{Social impact.}
Our method lowers the barrier to generating high-quality 3D facial avatars from a single image. Intended for authorized asset creation in film, games, virtual production, and telepresence, it could also enable non-consensual digital doubles, deepfake-ready identities, or impersonation assets from unauthorized images, harming privacy and identity security. It should therefore be used only with authorized images and valid usage rights. For released code and models, usage terms will prohibit impersonation, non-consensual identity replication, and other identity misuse; watermarking, metadata disclosure, and forgery-detection compatibility are feasible downstream mitigation options.

\section{Acknowledgments}
{\footnotesize\sloppy This work was supported by the National Key Research and Development Program of China (No. 2023YFC3306401), National Natural Science Foundation of China (No. 62576018), Zhejiang Provincial Natural Science Foundation of China (No. LD24F020007), Beijing Natural Science Foundation (No. L244043), and the Ministry of Economic Development of the Russian Federation (agreement 000000C313925P3X0002, grant No. 139-15-2025-004 dated 17.04.2025). We also thank Great Wall Motors for their support.\par}

\appendix
\setcounter{figure}{0}
\setcounter{table}{0}
\setcounter{equation}{0}
\renewcommand{\thefigure}{A\arabic{figure}}
\renewcommand{\thetable}{A\arabic{table}}
\renewcommand{\theequation}{A\arabic{equation}}

% \maketitle

\section{More Implementation Details}

\subsection{Differentiable GGX Microfacet Shader}
\label{sec:appendix_shading}

In this section, we provide the detailed mathematical formulation of our differentiable GGX microfacet shader $\mathcal{S}_{\mathrm{GGX}}$.
To achieve high fidelity texture synthesis while maintaining computational efficiency, our pipeline operates entirely in UV space.
Our implementation adheres to the physical-based rendering equation (PBR)~\cite{kajiya1986rendering} and incorporates specific numerical optimizations to ensure stable gradient propagation during the diffusion training process for all predicted components: albedo $\hat{T}_{\mathrm{alb}}$, normal $\hat{T}_{\mathrm{nrm}}$, roughness $\hat{T}_{\mathrm{rough}}$, specular $\hat{T}_{\mathrm{spec}}$ and displacement $\hat{T}_{\mathrm{disp}}$ maps.
Since our rendering process operates in UV space rather than relying on rasterization, we can employ fixed geometric priors.
Specifically, we adopt a fixed template face and use Blender to precompute its UV space world space position map $T_{\mathrm{pos}}$, geometric normal map $T^{\mathrm{geo}}_{\mathrm{nrm}}$, and geometric tangent map $T^{\mathrm{geo}}_{\mathrm{tan}}$.
These maps serve as geometric priors that assist the rendering, enabling us to calculate accurate 3D light transport directly in the 2D UV domain.

Before shading evaluation, we strictly define the geometric context.
First, the predicted displacement map $\hat{T}_{\mathrm{disp}}$ is applied along the geometric normal to update the pixel's world position:
\begin{equation}
    \mathbf{P}(\mathbf{u}) = \tilde{\mathbf{P}}(\mathbf{u}) + s_{\mathrm{disp}}\, \hat{T}_{\mathrm{disp}}(\mathbf{u})\, \mathbf{N}_g(\mathbf{u}),
\end{equation}
where $\mathbf{N}_g$ is the normalized version of $T^{\mathrm{geo}}_{\mathrm{nrm}}$, $s_{disp}$ is set 0.01.
Crucially, to maintain a consistent local shading frame aligned with the perturbed normal, we perform tangent space reorthogonalization.
Normal mapping introduces perturbations that misalign the original tangent frame with the shading normal. Specifically, the decoded tangent space normal
$\mathbf n_t=(n_x,n_y,n_z)$ is lifted to world space as
$\mathbf N_s=\mathrm{normalize}(n_x\mathbf T_g+n_y\mathbf B_g+n_z\mathbf N_g)$,
where $\{\mathbf T_g,\mathbf B_g,\mathbf N_g\}$ denotes the geometric tangent frame with $\mathbf B_g=\mathrm{normalize}(\mathbf N_g\times \mathbf T_g)$.
Subsequently, we use the Gram-Schmidt process to construct a corrected local tangent $\mathbf{T}_\ell$ that is enforced to be orthogonal to $\mathbf{N}_s$:
\begin{equation}
    \mathbf{T}_\ell = \mathrm{normalize}\!\left( T^{\mathrm{geo}}_{\mathrm{tan}} - (T^{\mathrm{geo}}_{\mathrm{tan}} \cdot \mathbf{N}_s)\mathbf{N}_s \right).
\end{equation}
The bitangent is derived as $\mathbf{B}_\ell = \mathrm{normalize}(\mathbf{N}_s \times \mathbf{T}_\ell)$. All view vectors $\mathbf{v}$ and light vectors $\mathbf{l}$ are projected into this corrected local basis $\{\mathbf{T}_\ell, \mathbf{B}_\ell, \mathbf{N}_s\}$ for subsequent BRDF evaluation.

Regarding material modeling, we employ the Cook-Torrance microfacet model~\cite{cook1982reflectance} with a metallic-roughness parameterization.
We assume the skin is a non-metallic material and use the predicted specular map $\hat{T}_{\mathrm{spec}}$ to modulate the Fresnel base reflectance $F_0 = 0.08 \cdot \hat{T}_{\mathrm{spec}}$.
The diffuse term uses the Lambertian model $f_d = \hat{T}_{\mathrm{alb}} / \pi$.
For the specular term, we use the GGX distribution function $D_{\mathrm{GGX}}$.
To improve numerical stability at grazing angles, we adopt a simplified GGX formulation
in which the classical normalization denominator is omitted and set to a constant value.
This yields a bounded, differentiable shading proxy that does not strictly enforce energy conservation:
\begin{equation}
    f_s(\mathbf{u},\mathbf{v},\mathbf{l}) = D_{\mathrm{GGX}}(\mathbf{n} \cdot \mathbf{h}, \alpha^2) \, F(\mathbf{l} \cdot \mathbf{h}) \, V_{\mathrm{Smith}}(\mathbf{n} \cdot \mathbf{l}, \mathbf{n} \cdot \mathbf{v}, \alpha^2),
\end{equation}
where $D_{\mathrm{GGX}}$ is the Trowbridge-Reitz distribution, and $F$ is the Schlick approximation.
The visibility term $V_{\mathrm{Smith}}$, which analytically combines the geometry masking term and the BRDF normalization factor, is formulated as:
\begin{equation}
    V_{\mathrm{Smith}} = \frac{0.5}{(\mathbf{n} \cdot \mathbf{v}) \sqrt{\alpha^2 + (1-\alpha^2)(\mathbf{n} \cdot \mathbf{l})^2} + (\mathbf{n} \cdot \mathbf{l}) \sqrt{\alpha^2 + (1-\alpha^2)(\mathbf{n} \cdot \mathbf{v})^2}}.
\end{equation}
Here, $\alpha = \mathrm{roughness}^2$, and the terms $\mathbf{n} \cdot \mathbf{l}$ and $\mathbf{n} \cdot \mathbf{v}$ denote the clamped dot products in the local shading frame.

\begin{figure}[t]
  \centering
  \includegraphics[width=0.9\linewidth]{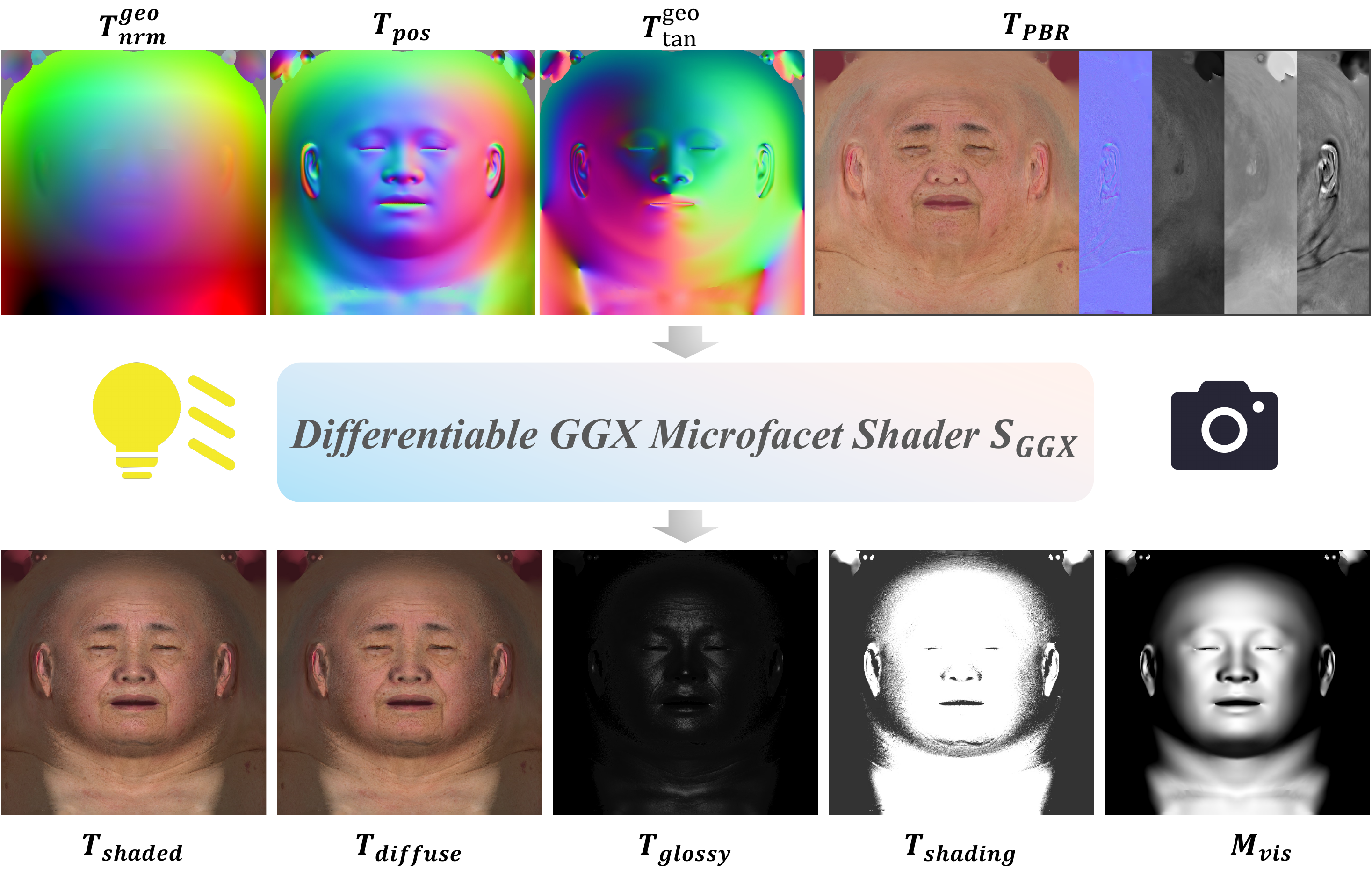}
  \caption{\textbf{ Decomposition of Physically Based Shading Components in UV Space.}
    The figure illustrates how decoupled geometric priors and predicted PBR maps are processed by our differentiable shader to yield final rendering results and their constituent components.
    Top Row (Inputs): The geometric context is established by the world space geometric normal map $\boldsymbol{T}^{\mathrm{geo}}_{\mathrm{nrm}}$, position map $\boldsymbol{T}_{\mathrm{pos}}$, and geometric tangent map $\boldsymbol{T}^{\mathrm{geo}}_{\mathrm{tan}}$. These are combined with the predicted material parameter maps $\boldsymbol{T}_{\mathrm{PBR}}$ (containing albedo, normal, roughness, specular, etc.).
    Middle: The differentiable GGX microfacet shader $S_{\mathrm{GGX}}$ integrates these inputs under specific lighting and view conditions.
    Bottom Row (Outputs \& Components): The results show the final composite shaded texture $\boldsymbol{T}_{\mathrm{shaded}}$, and its decomposition into the diffuse component $\boldsymbol{T}_{\mathrm{diffuse}}$, the specular (glossy) component $\boldsymbol{T}_{\mathrm{glossy}}$, and the pure illumination intensity map $\boldsymbol{T}_{\mathrm{shading}}$. The geometric visibility mask $\boldsymbol{M}_{\mathrm{vis}}$, used for loss calculation, is also shown.}
  \label{supfig:shader}
\end{figure}

The final rendering equation evaluates direct illumination by accumulating contributions from a set of sampled directional light directions.
To prevent vanishing gradients in fully occluded regions (where direct lighting is blocked), we combine the direct illumination with a simplified ambient term:
\begin{equation}
\label{eq:final_rendering}
\begin{split}
\hat{T}_{\mathrm{shaded}}(\mathbf{u})
=
M(\mathbf{u})
\bigg[
& \sum_{k=1}^{K}
L_k(\mathbf{l}_k)\,
(\mathbf{n} \cdot \mathbf{l}_k)^+\,
\Big(
f_d(\mathbf{u}) + f_s(\mathbf{u},\mathbf{v},\mathbf{l}_k)
\Big)
\\
& +
I_{\mathrm{amb}} \cdot \hat{T}_{\mathrm{alb}}(\mathbf{u})
\bigg],
\end{split}
\end{equation}
where $L_k(\mathbf l_k)$ denotes the radiance returned by the lighting module for the $k$-th sampled light direction, and $(\mathbf{n} \cdot \mathbf{l}_k)^+ = \max(0, \mathbf{n} \cdot \mathbf{l}_k)$ denotes the clamped cosine term using the shading normal.
The ambient intensity $I_{\mathrm{amb}}$ is randomly sampled from a uniform distribution $\mathcal{U}(0.15, 0.3)$ during training.
For supervision, we additionally compute a visibility weight
\begin{equation}
M_{\mathrm{vis}}(\mathbf{u})
=
M(\mathbf{u})\cdot
\frac{1}{K}\sum_{k=1}^{K}
(\mathbf{N}_g(\mathbf{u})\cdot \mathbf{v}(\mathbf{u}))^+,
\end{equation}
which is used only for loss weighting and is not applied to the forward shading result.

All shading computations are performed in linear color space. 
Input albedo maps are converted from sRGB to linear space prior to shading. 
For visualization and loss evaluation, rendered images are converted back to gamma space.
For clarity, gamma transformations are omitted in the mathematical formulation above.
Fig.~\ref{supfig:shader} provides an overview of the differentiable shading pipeline and its constituent components.

\subsection{Details of LoRA Injected Layers}
\label{app:lora_layers}

The LoRA adapters are specifically injected into the following attention and feedforward projection layers of the unified diffusion transformer (DiT) backbone:
\begin{itemize}
    \item \texttt{to\_k}, \texttt{to\_q}, \texttt{to\_v}
    \item \texttt{to\_out.0}
    \item \texttt{add\_k\_proj}, \texttt{add\_q\_proj}, \texttt{add\_v\_proj}
    \item \texttt{to\_add\_out}
    \item \texttt{ff.net.0.proj}, \texttt{ff.net.2}
    \item \texttt{ff\_context.net.0.proj}, \texttt{ff\_context.net.2}
\end{itemize}

\section{More Dataset Preparation Details}
\label{supp:dataset_prep}

To train our high fidelity avatar reconstruction pipeline, we meticulously constructed paired data tailored for each stage of the system.
Our core strategy involves synergizing a limited amount of high quality 3D texture data with existing large scale public 2D face datasets.
Specifically, we leverage two large scale public datasets, FFHQ~\cite{karras2019style} and CelebAMask-HQ~\cite{CelebAMask-HQ}, collectively containing nearly 100,000 in-the-wild facial images (denoted as $I_{wild}$) spanning wide variations in ethnicity, age, expression, pose, and occlusion.
We employ these datasets to train our geometry reconstruction model and to provide occlusion priors for the texture generation pipeline.
Regarding texture and material learning, unlike prior methods that rely on large scale GAN generated synthetic textures~\cite{li2024uv,bai2023ffhq} or expensive multi view capture systems~\cite{Lattas2022AvatarMe++,fitdiff,Lattas2023fitme,lattas2020avatarme,Papantoniou2023Relightify}, we utilize a compact yet high quality dataset $D_{scan}$\footnote{https://www.3dscanstore.com.}, containing fewer than 100 professional 3D face scans obtained online.
Each scan provides a complete set of physically based rendering (PBR) texture maps ($T_{PBR}$) at resolutions up to 8K, including albedo $T_{alb}$, normal $T_{nrm}$, specular $T_{spec}$, roughness $T_{rough}$, and displacement maps $T_{disp}$.
Rather than seeking comprehensive appearance diversity, we treat these scans as physically accurate "anchors."
Our core insight is that combining such high fidelity ground truth with the priors of modern pretrained image generation diffusion models enables accurate and generalizable material estimation even with limited scan data.

% 移除 \paragraphspace，直接使用 paragraph
\paragraph{Large Scale Synthetic 3D Dataset Construction.}
As illustrated in Fig.~3, to amplify the utility of limited groundtruth material data,
we construct a large scale synthetic 3D face dataset.
Using our geometry reconstruction network (Sec. 3.2),
we extract 3DMM coefficients from $I_{wild}$ to reconstruct diverse facial geometries $\mathbf{G}$.
Subsequently, high quality PBR textures $T_{PBR}$ from $D_{scan}$ are randomly assigned to these geometries,
generating 100,000 synthetic instances.
To simulate realistic data incompleteness, we employ the DMLCSR segmentation model~\cite{dml} to extract visible skin regions $I_{vis}$.
Leveraging the geometry based UV mapping, we unwrap these regions into UV space to compute the visibility mask $M_{vis}$,
which explicitly encodes both viewpoint dependent self occlusions and external occlusions caused by hair, glasses, or accessories.

\paragraph{Texture Inpainting Data.}
The goal of the texture inpainting module is to recover complete UV textures from partial and occluded observations.
To generate realistic training pairs, we render the synthetic 3D models using the Blender Cycles engine under 2,041 diverse HDRI environment maps ($L_{env}$)~\cite{li2025lino}, with additional rotation augmentation.
This process yields photorealistic shaded images $I_{env}$.
Subsequently, the full shading effects under the specific viewpoint and illumination are baked back into UV space, generating complete shaded textures $T_{env}$ that serve as the groundtruth targets.
To construct the incomplete input texture $T_{inc}$, rather than simply applying a mask to $T_{env}$, we reunwrap the rendered image $I_{env}$ back into UV space and multiply it by the pre-computed visibility mask $M_{vis}$, thereby incorporating inherent occlusions and interpolation artifacts.

\paragraph{Light Homogenization Data.}
The light homogenization module aims to remove scene specific illumination effects.
Its input is the shaded texture $T_{env}$.
Following the formulation in NEAR~\cite{li2025near} and MoSAR~\cite{dib2024mosar},
we render and bake the same instance using uniform all white ambient lighting $L_{uni}$
to obtain the light-homogenized texture $T_{hom}$.
This process preserves intrinsic material cues
while normalizing all samples into a unified lighting environment,
significantly reducing the difficulty of subsequent intrinsic decomposition.

\paragraph{Intrinsic Material Estimation Data.}
Given the light-homogenized texture $T_{hom}$ as input,
the intrinsic material estimation module directly predicts physically meaningful material properties.
Since lighting has been explicitly normalized,
we use the corresponding groundtruth PBR texture set $T_{PBR}$
as supervision for this stage.

\paragraph{Super Resolution Data.}
Since the preceding texture data are processed at 1K resolution for efficiency, we downsample the 8K textures to 4K to serve as the super resolution ground truth, thereby meeting the demands of high fidelity rendering.

\begin{table}[t]
\small
\centering
\caption{Inference latency breakdown of our multi stage pipeline. All stages are evaluated on a single NVIDIA H100 GPU for a 1K resolution input (upscaled to 4K in the final stage). Sampling steps are set to 30 for all diffusion based modules with a guidance scale of 2.0.}
\label{tab:latency}
\resizebox{\linewidth}{!}{%
\begin{tabular}{lllc}
\toprule
\textbf{Stage} & \textbf{Backbone} & \textbf{Time} \\ \midrule
Geometry Reconstruction & ConvNeXt V2 + DINOv3 & $<$ 0.5s \\
Texture Inpainting & Flow matching DiT + LoRA & 30s \\
Light Homogenization  & Flow matching DiT + LoRA  & 30s \\
Intrinsic Material Estimation &  Joint Diffusion w/ Cross Attention  & 3 min \\
Super Resolution & RealESRGAN (1K $\rightarrow$ 4K) & $\sim$ 2s \\ \midrule
\textbf{Total Latency} & & \textbf{$\sim$ 4 min} \\ \bottomrule
\end{tabular}%
}
\end{table}

\begin{table}[htbp]
\centering
\caption{User Study results. Values represent the preference rate (\%) of our method against baselines across three dimensions.}
\label{tab:user_study}
\resizebox{\linewidth}{!}{%
\begin{tabular}{@{}lccc@{}}
\toprule
\textbf{Method} & \textbf{Geometric Details} & \textbf{Texture Realism} & \textbf{Relighting Quality} \\ \midrule
vs. MoSAR \cite{dib2024mosar} & \textbf{60.0\%} & \textbf{83.3\%} & \textbf{80.0\%} \\
vs. FitMe \cite{Lattas2023fitme} & \textbf{96.7\%} & \textbf{93.3\%} & \textbf{93.3\%} \\ 
vs. Relightify \cite{Papantoniou2023Relightify} & \textbf{100.0\%} & \textbf{100.0\%} & \textbf{100.0\%} \\
\bottomrule
\end{tabular}%
}
\end{table}

\section{User Study}
To evaluate avatar reconstruction quality, we conducted a user study with 30 participants and 20 reconstruction sets. Since MoSAR \cite{dib2024mosar}, FitMe \cite{Lattas2023fitme}, and Relightify \cite{Papantoniou2023Relightify} are not open sourced, we utilized high resolution results from MoSAR’s official supplementary materials for a fair comparison. Participants evaluated methods across three dimensions: geometric precision, texture realism (covering age, ethnicity, and occlusions), and relighting quality.

As shown in Tab.~\ref{tab:user_study}, our method was significantly preferred across all categories. Notably, we achieved a 100.0\% preference rate over Relightify and consistently outperformed FitMe in both geometric details (96.7\%) and relighting quality (93.3\%). While MoSAR showed competitive geometric performance, our method still maintained a clear advantage in texture realism (83.3\%) and relighting quality (80.0\%), demonstrating superior robustness in handling complex in-the-wild lighting and fine grained material recovery.

\section{More Visual Results}
In this section, we provide extensive qualitative results to further demonstrate the robustness and fidelity of our proposed framework across diverse and challenging scenarios.

\begin{figure}[t]
    \centering
    \includegraphics[width=1\linewidth]{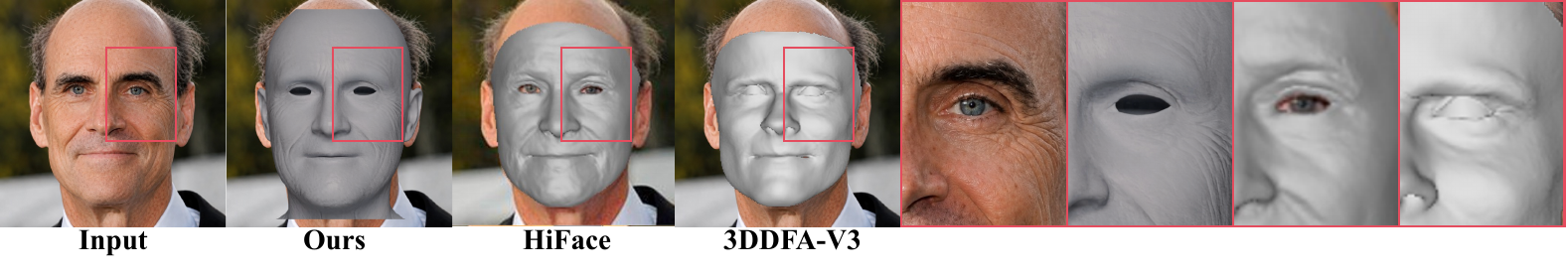}
    \caption{
    \textbf{Additional geometry comparison.}
    By leveraging predicted normal and displacement maps, our method recovers clearer facial high-frequency geometric details, such as crow's feet, compared with HiFace and 3DDFA-V3.
    }
    \label{suppfig:rebuttal_geo}
\end{figure}

\begin{figure}[t]
    \centering
    \includegraphics[width=1\linewidth]{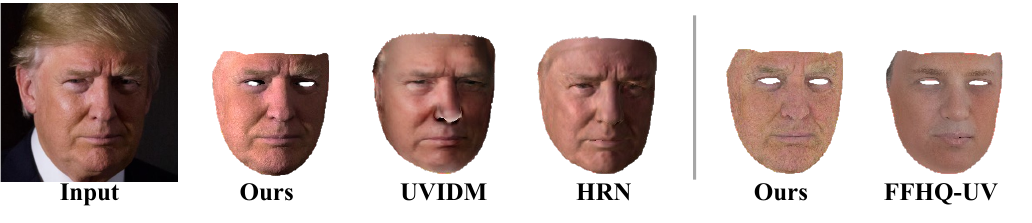}
    \caption{
    \textbf{Additional texture inpainting and light homogenization comparison.}
    Left: compared with UV-IDM and HRN, our inpainting produces cleaner UV textures with fewer baked-in occlusions and better identity preservation.
    Right: compared with FFHQ-UV, our light homogenization yields a more uniformly lit texture while retaining high-frequency skin details.
    }
    \label{suppfig:rebuttal_tex}
\end{figure}

\paragraph{Geometric Fidelity and Micro-structures.} 
In Figs.~\ref{suppfig:rebuttal_geo},~\ref{suppfig:geo1} and~\ref{suppfig:geo2}, we compare our displaced geometry, which applies the generated normal and displacement maps to the estimated mesh, against a series of representative single-image reconstruction algorithms. In particular, Fig.~\ref{suppfig:rebuttal_geo} highlights that our predicted normal and displacement maps recover clearer facial high-frequency geometric details, such as crow's feet, compared with HiFace and 3DDFA-V3. 3DDFA-V3~\cite{wang20243d} and Deep3D~\cite{deng2019accurate} tend to produce smooth, generic facial shapes. Although they can capture the overall contours, they lose subtle details that reflect individual identity characteristics (such as skin texture and shallow wrinkles), and 3DDFA-V3 additionally suffers from excessive geometric deformation due to segmentation clustering. DECA~\cite{DECA} and EMOCA~\cite{EMOCA} attempt to estimate additional displacement maps to represent facial details, but this often leads to distorted geometric artifacts and wavy shading normals. While SMIRK~\cite{SMIRK} demonstrates significant improvements in expression estimation, it struggles to generate physically plausible and fine geometric details. HRN~\cite{lei2023hierarchical} attempts to enhance fidelity by forcibly baking image details into the geometry, but this typically causes severe topological artifacts. In contrast, our approach utilizes predicted high-precision normal and displacement maps to synergistically drive surface deformation. This decoupled method allows us to recover highly challenging micro-structures (e.g., deep wrinkles and pore details in elderly individuals) while maintaining a clean and continuous geometric topology.

\paragraph{Texture Inpainting and Light-homogenization.}
Fig.~\ref{suppfig:rebuttal_tex} provides additional qualitative comparisons for the texture completion and illumination disentanglement stages. Compared with UV-IDM and HRN, our inpainting module produces cleaner UV textures with fewer baked-in occlusions and better identity preservation. Compared with FFHQ-UV, our light-homogenization yields a more uniformly lit albedo while retaining high-frequency skin details, which benefits the subsequent PBR material estimation.

\paragraph{Generalization on In-the-wild Inputs.}
Fig.~\ref{suppfig:ours1},~\ref{suppfig:ours2} and~\ref{suppfig:ours3} demonstrate that despite significant variations in ethnicity, age, and initial lighting conditions, our pipeline consistently achieves accurate intrinsic decomposition and photorealistic relighting results. These extended results highlight the remarkable generalization capabilities and high practical utility of our data efficient diffusion based approach, even when trained on limited professional scan data.

\paragraph{Robustness to Challenging Poses and Illumination.} Figs.~\ref{suppfig:ours4} and~\ref{suppfig:ours5} specifically showcase our model's performance on images with extreme head poses and complex lighting environments.
\begin{itemize}
    \item \textbf{Extreme Pose Handling:} In cases of extreme yaw and pitch, such as the profile and upward facing views in Figs.~\ref{suppfig:ours4} and~\ref{suppfig:ours5}, our pipeline maintains identity consistency and geometric accuracy. The combination of our geometry encoder and the diffusion based texture inpainting module successfully hallucinates missing information in self occluded regions while preserving high frequency details on visible surfaces.
    \item \textbf{Challenging Illumination:} Our light-homogenization module demonstrates a remarkable capability in disentangling challenging illumination. Even under conditions with strong cast shadows or high contrast directional lighting, such as the samples in the bottom rows of Fig.~\ref{suppfig:ours4} and~\ref{suppfig:ours5}, our method effectively removes these scene specific artifacts to produce clean, uniform albedo and material maps.
\end{itemize}

\begin{figure}[t]
    \centering
    \includegraphics[width=1\linewidth]{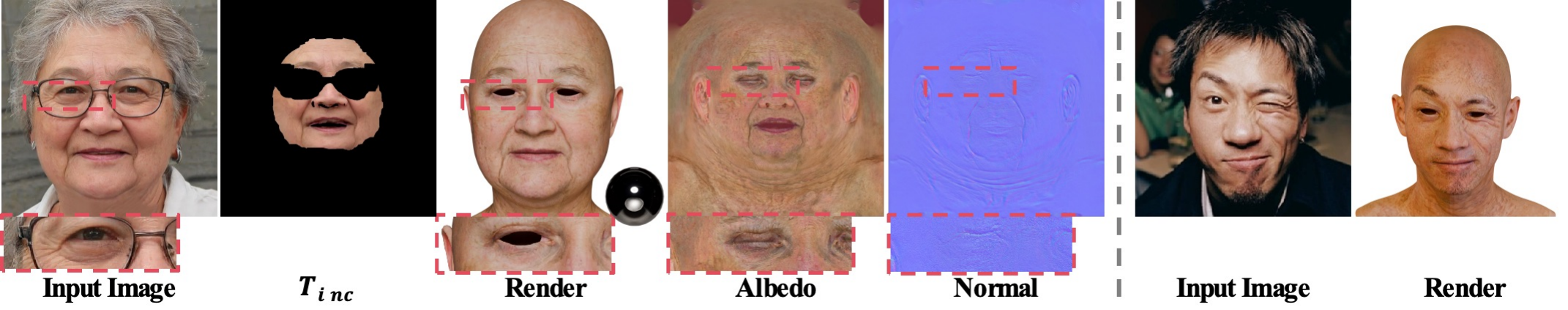}
    \caption{
    \textbf{
    Limitations in handling semi transparent occlusions and extreme expressions.} (Left) While our method successfully use pretrained face segmentation model~\cite{dml} removes eyeglasses, the restored eye region may appear over-smoothed as the model relies on general generative priors due to the lack of visible high frequency information (e.g., specific crow's feet) in the incomplete texture $T_{inc}$. (Right) Challenges under extreme expressions: our current pipeline relies on the Hifi3D++ morphable model for initial geometry. As shown in the winking example, when the linear 3DMM basis fails to accurately capture highly non-rigid deformations (e.g., asymmetric squinting), the resulting rendering may exhibit geometric misalignment or loss of identity specific details. However, from the rendering results, our method still preserves high-fidelity local skin structures, such as eyebrows and wrinkles. 
    }
    \label{fig:limitations}
\end{figure}

\section{Limitations}
Our method has several inherent limitations that we aim to address in future iterations. 

\paragraph{Occlusions and Detail Loss.} While our framework is robust to general occlusions like hair, recovering fine grained details behind semi transparent materials remains challenging. As illustrated in Fig.~\ref{fig:limitations} (Left), although we employ a third party segmentation model to remove eyeglasses, the restored eye region often lacks user specific high frequency details, such as crow's feet, because the model relies on general generative priors when original information is obstructed. We plan to augment our synthetic dataset with diverse 3D accessories to force the model to "see behind" occlusions during training without external segmentation.

\paragraph{Geometric Basis and Extreme Expressiveness.} A significant constraint arises from our choice of the linear Hifi3D++~\cite{bao2021high} 3DMM basis for initial geometry reconstruction. This basis struggles to model extreme facial expressions or highly non-rigid deformations. As shown in Fig.~\ref{fig:limitations} (Right), when the geometric fitting fails to accurately capture asymmetric expressions like a wink, it results in misaligned surface details or local artifacts in the rendered output, despite the PBR textures maintaining local fidelity. Future work will investigate nonlinear geometric representations to ensure higher quality reconstruction and animation.

\paragraph{Inference Latency.} Our multi stage pipeline introduces significant inference latency compared to end-to-end regression networks. As detailed in Tab.~\ref{tab:latency}, on a single H100 GPU (30 steps, guidance scale 2.0), inpainting and light homogenization each take approximately 30 seconds, while joint multi branch material estimation requires about 3 minutes. While this is suitable for high quality offline asset creation, we are exploring distillation and acceleration strategies, such as consistency, adversarial or distribution matching distillation \cite{Consistency, ADD, jiang2025distribution, team2025zimage}, to reduce this latency without compromising quality.

\paragraph{Editability and Priors.} Lastly, finetuning LoRA adapters on a compact dataset of professional 3D scans tends to degrade the open domain text editing capabilities of the base diffusion model. Maintaining a balance between physically accurate material estimation and flexible text based appearance modification remains an ongoing challenge for future research.

\begin{figure}[t]
    \centering
    \includegraphics[width=1\linewidth]{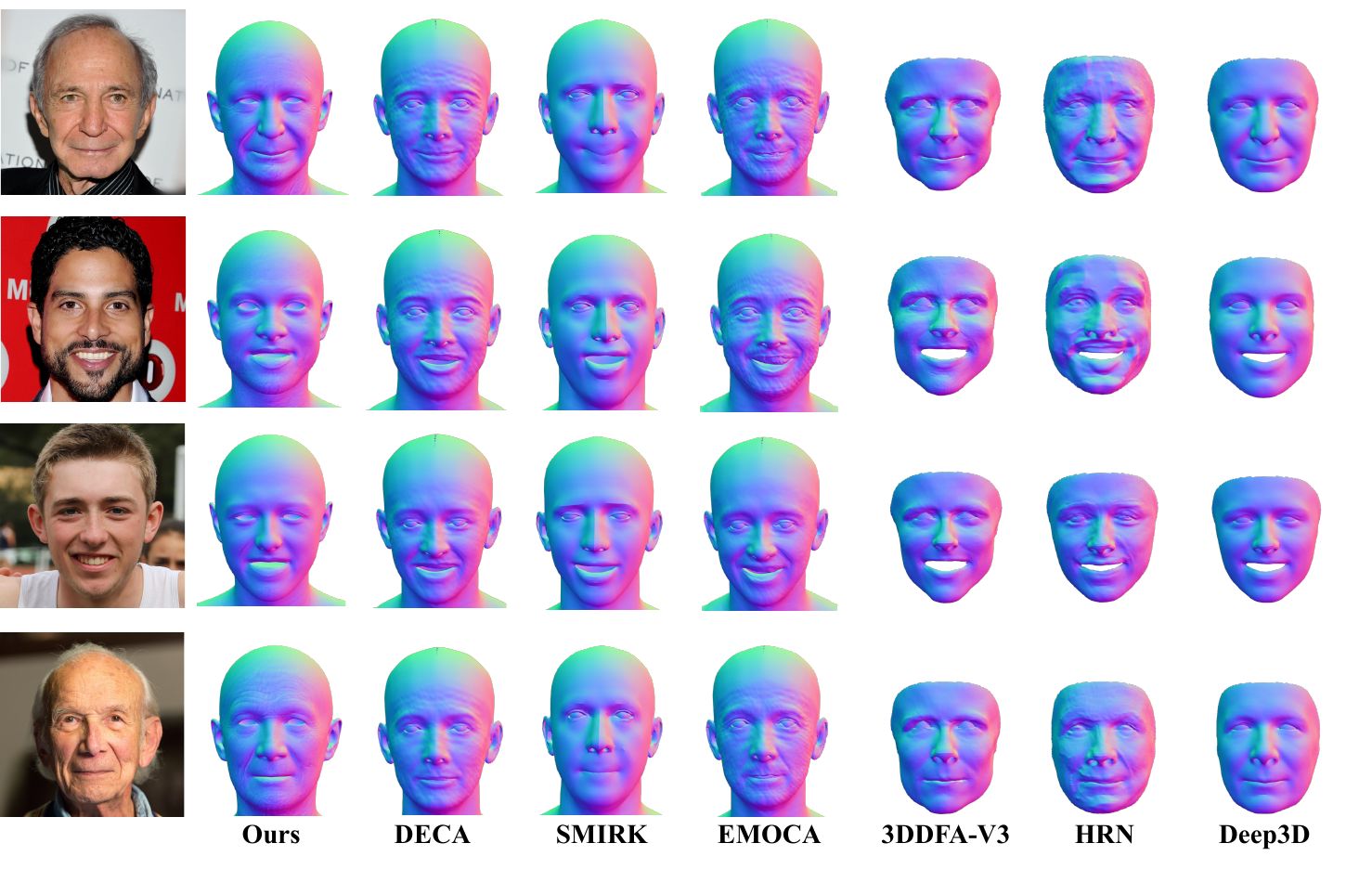}
    \caption{\textbf{Qualitative comparison of geometric fidelity.} 
    We compare our displaced geometry against various state-of-the-art methods. 
    While DECA and EMOCA attempt to reconstruct faces via displacement estimation, they often introduce distorted facial details and artifacts, such as the ripple-like noise in the normal maps (Columns 3 and 5). 
    Frameworks like SMIRK and Deep3D yield overly smoothed facial shapes, whereas HRN tends to bake illumination-dependent details (e.g., shadows or specular highlights) into the geometry, resulting in severe high-frequency artifacts, such as the unnaturally flat regions on the forehead (Row 2). 
    Furthermore, 3DDFA-V3 fails to represent intricate details like expression-dependent wrinkles and smile lines, often leading to excessive geometric distortions in regions such as the brow ridge. 
    In contrast, our method achieves superior geometric fidelity by leveraging predicted Normal and Displacement maps to drive surface deformation. 
    This approach enables the recovery of complex identity-specific micro-structures (e.g., deep wrinkles and pores) that are geometrically precise yet topologically clean.
    }
    \label{suppfig:geo1}
\end{figure}

\begin{figure}[t]
    \centering
    \includegraphics[width=1\linewidth]{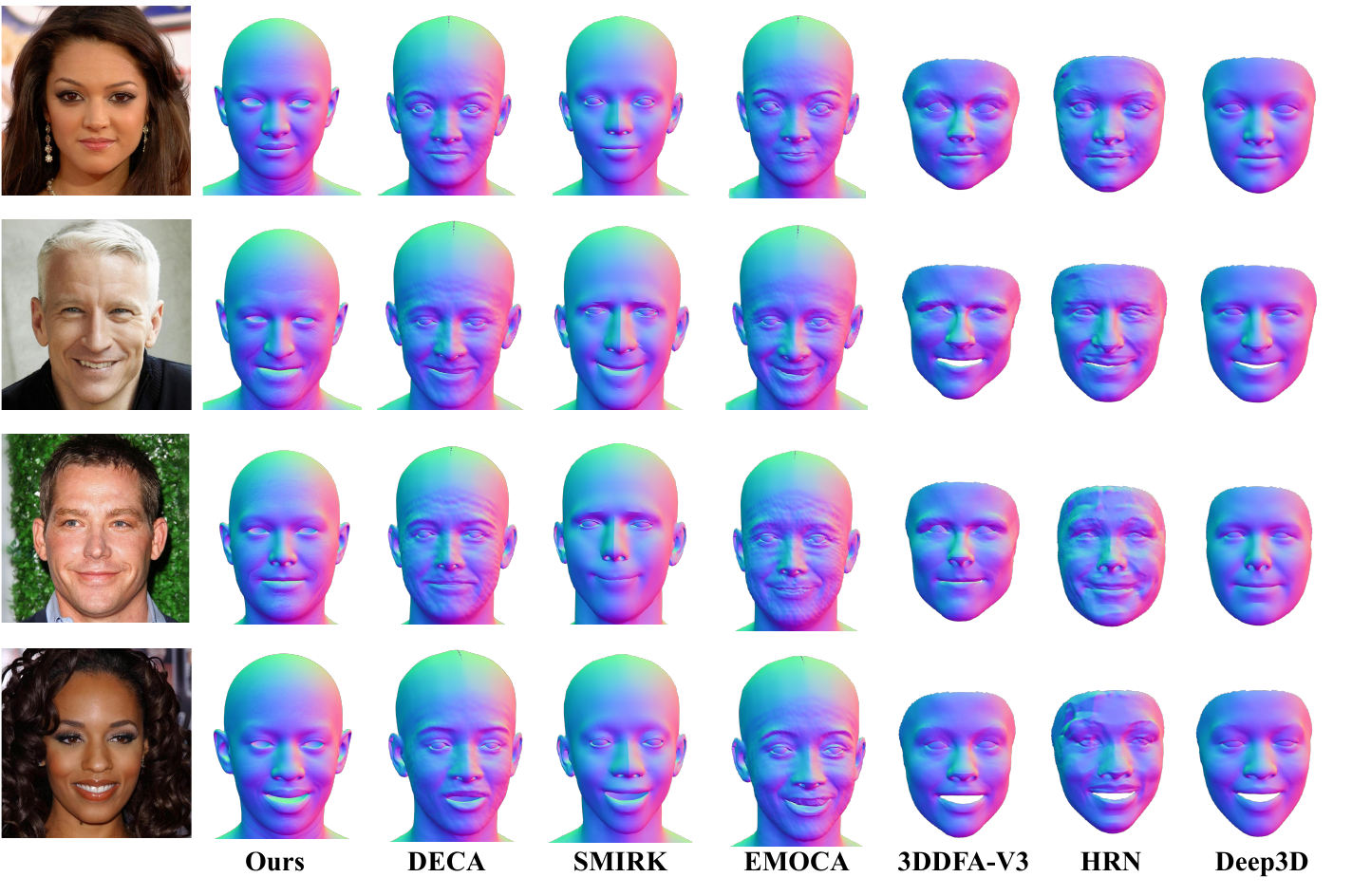}
    \caption{
    \textbf{Qualitative comparison of geometric fidelity.} Compared to alternative geometric estimation methods, our approach leverages normal and displacement maps to achieve significantly sharper facial details and superior realism in shading normal.
    }
    \label{suppfig:geo2}
\end{figure}

\begin{figure}[t]
    \centering
    \includegraphics[width=1\linewidth]{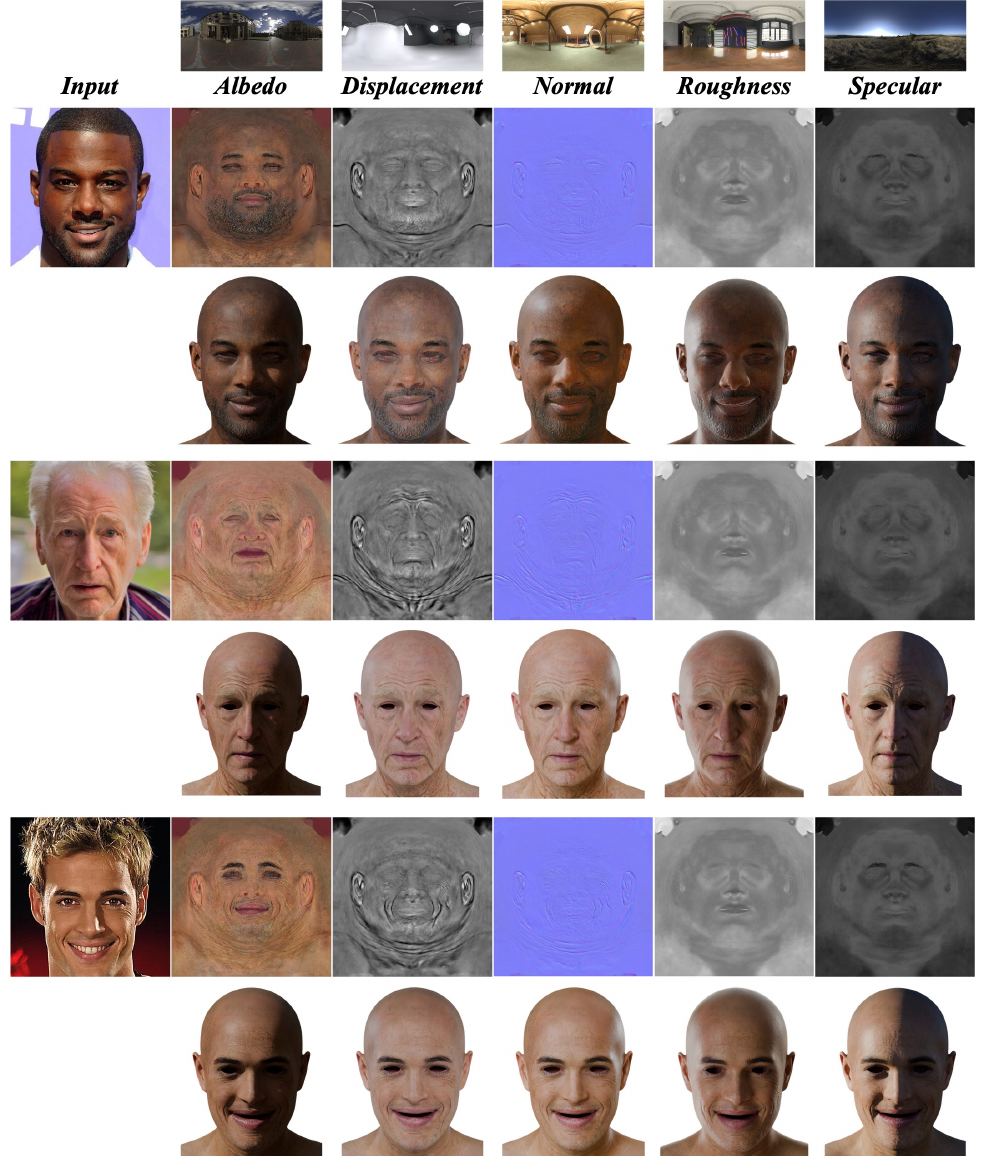}
    % \vspace{-30pt}
    \caption{
    For "in-the-wild" images, our pipeline robustly infers across different ethnicities, ages, and genders. This inference produces complete PBR textures and relighting renderings under a variety of environmental lighting conditions.
    }
    \label{suppfig:ours1}
\end{figure}

\begin{figure}[t]
    \centering
    \includegraphics[width=1\linewidth]{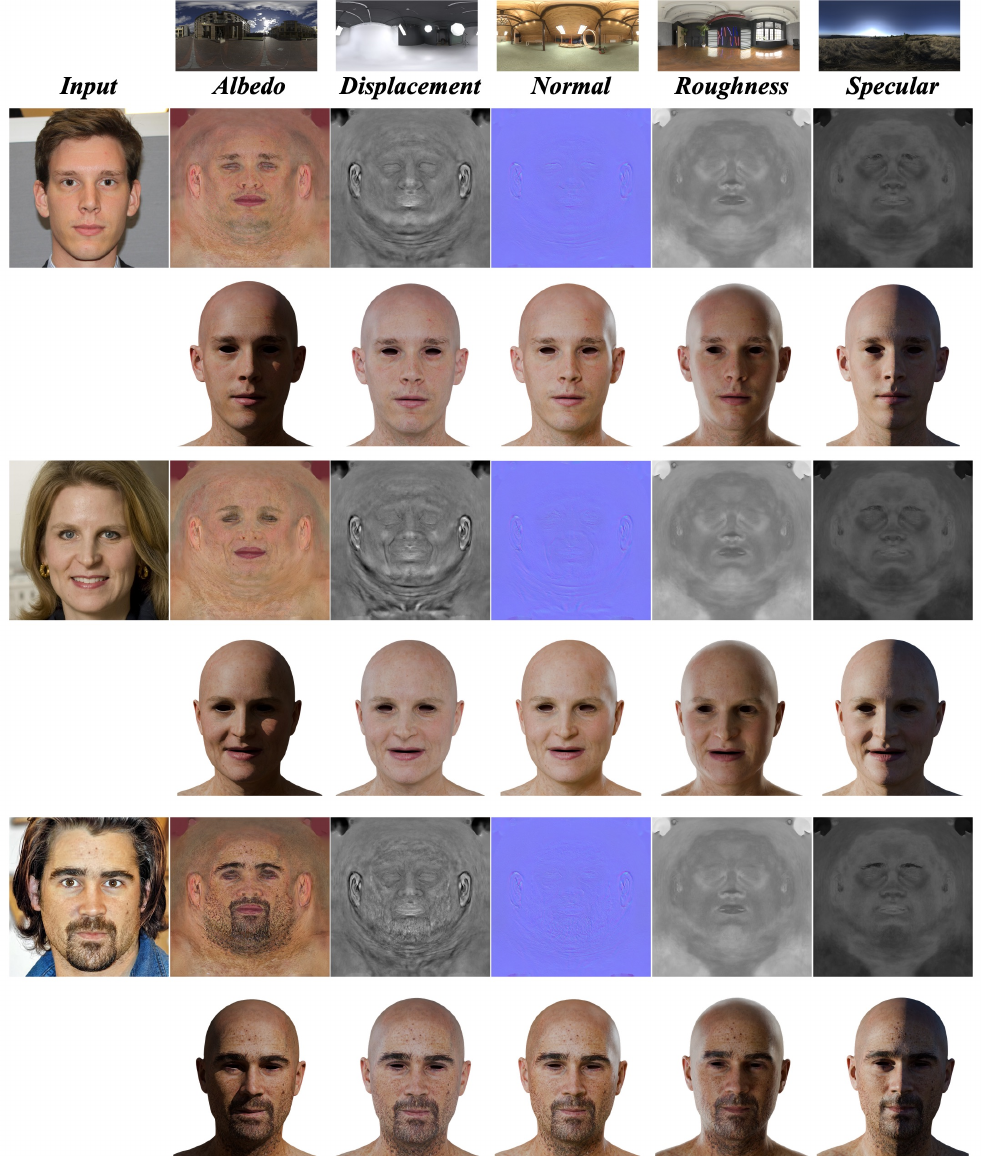}
    % \vspace{-30pt}
    \caption{
    For "in-the-wild" images, our pipeline robustly infers across different ethnicities, ages, and genders. This inference produces complete PBR textures and relighting renderings under a variety of environmental lighting conditions.
    }
    \label{suppfig:ours2}
\end{figure}

\begin{figure}[t]
    \centering
    \includegraphics[width=1\linewidth]{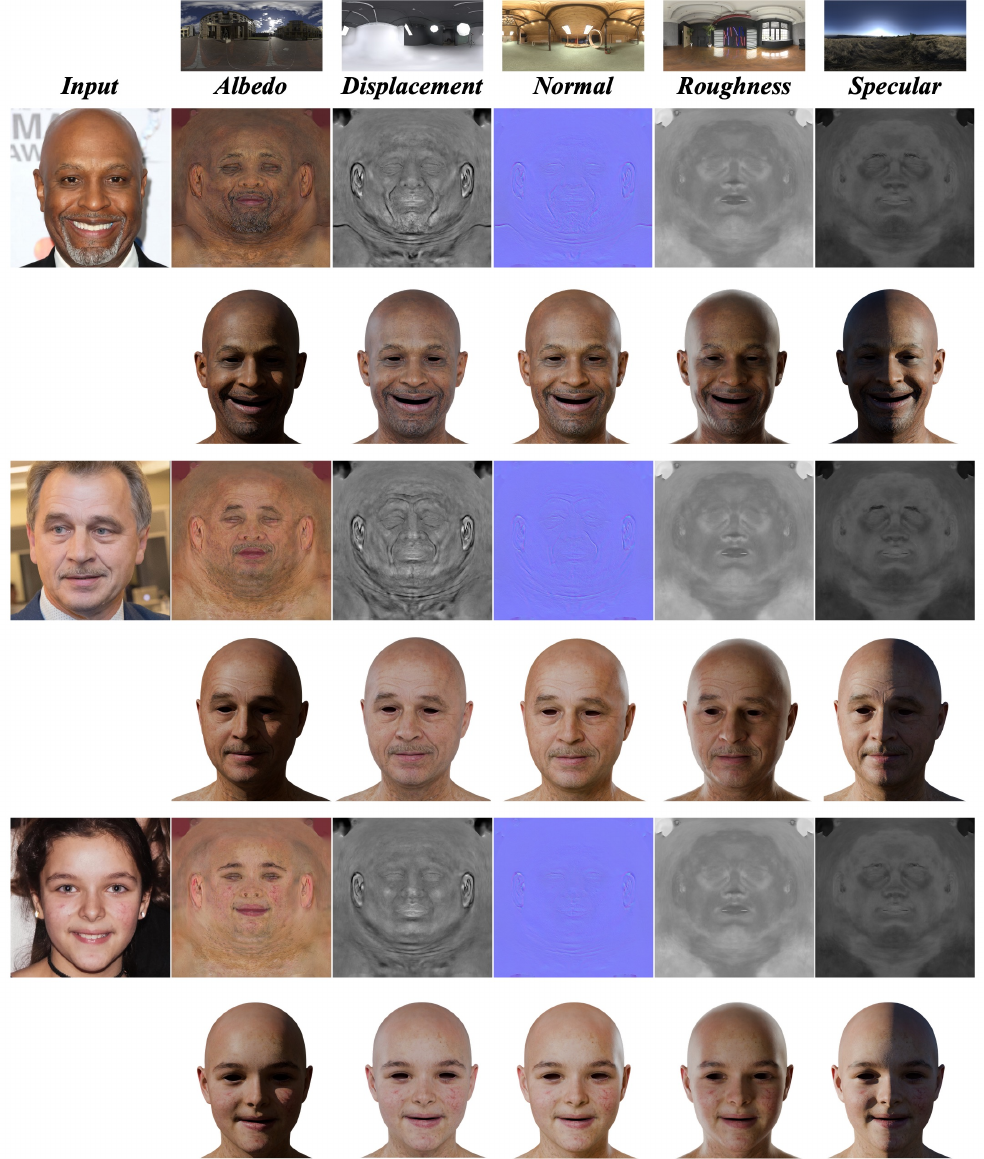}
    % \vspace{-30pt}
    \caption{
    For "in-the-wild" images, our pipeline robustly infers across different ethnicities, ages, and genders. This inference produces complete PBR textures and relighting renderings under a variety of environmental lighting conditions.
    }
    \label{suppfig:ours3}
\end{figure}

\begin{figure}[t]
    \centering
    \includegraphics[width=1\linewidth]{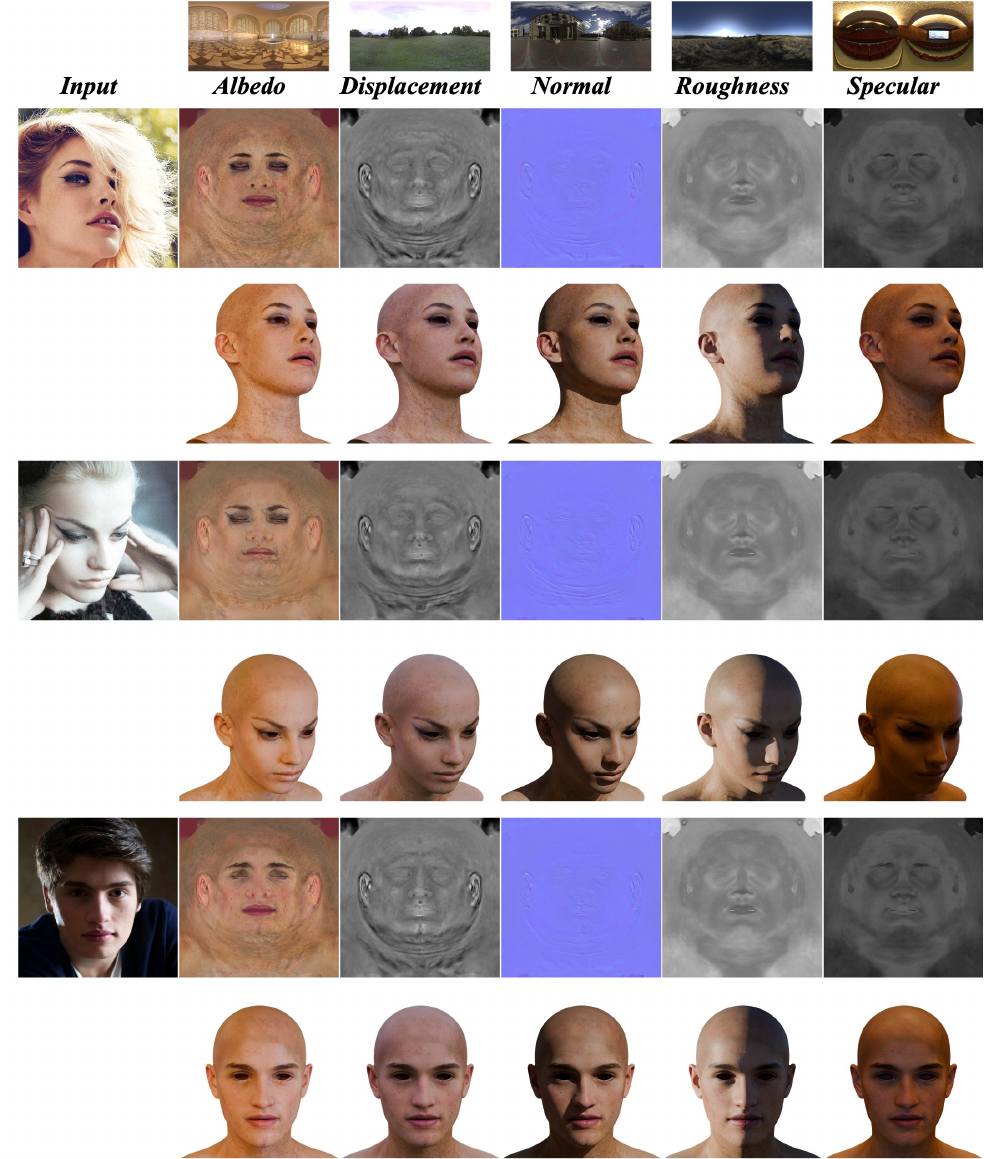}
    % \vspace{-30pt}
    \caption{\textbf{Robustness to extreme poses and challenging illumination.} 
    Our pipeline demonstrates significant robustness against large yaw and pitch angles, effectively recovering missing facial textures caused by severe self occlusions (e.g., hair or hands) while maintaining identity consistency, as shown in the first and second rows. In the third row, our method successfully disentangles intrinsic material properties even from inputs with high contrast cast shadows. The corresponding PBR components and relighting results under diverse HDRIs further demonstrate the model's capability to preserve high fidelity surface micro-structures.}
    \label{suppfig:ours4}
\end{figure}

\begin{figure}[t]
    \centering
    \includegraphics[width=1\linewidth]{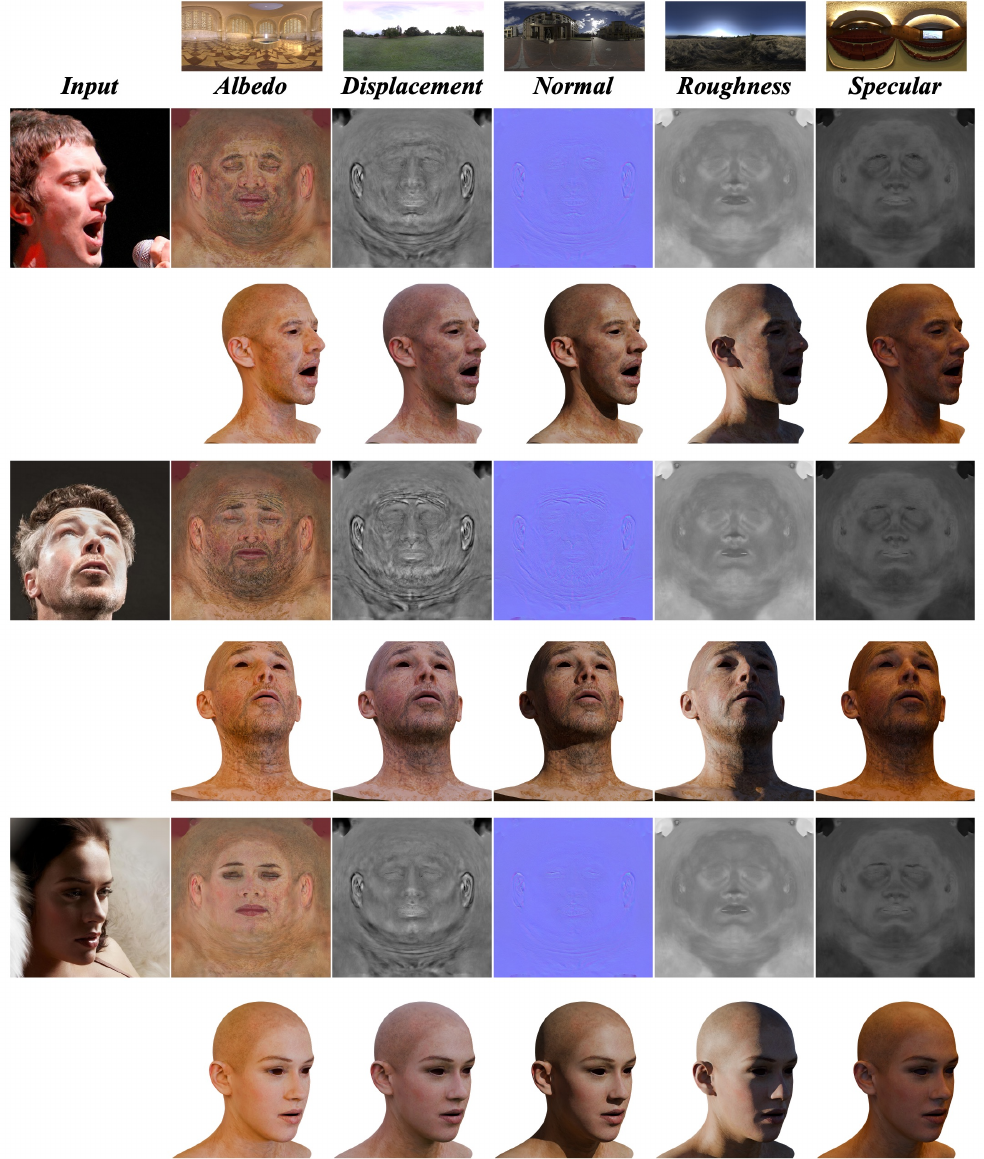}
    % \vspace{-30pt}
    \caption{
    \textbf{Robustness to extreme poses and challenging illumination.}
    Our method demonstrates significant robustness in handling large yaw and pitch angles. Even under complex conditions such as colored illumination (Row 1), over exposure (Row 2), and high contrast cast shadows (Row 3), our pipeline accurately disentangles intrinsic material properties.
    }
    \label{suppfig:ours5}
\end{figure}

% 插入参考文献（如果需要单独的文献列表）
\bibliographystyle{splncs04}
\bibliography{main}

% \section*{Acknowledgements}
% Please insert your acknowledgments here.
\clearpage
% ---- Bibliography ----
%
% BibTeX users should specify bibliography style 'splncs04'.
% References will then be sorted and formatted in the correct style.
%
\bibliographystyle{splncs04}
\bibliography{main}

\begin{thebibliography}{10}
\providecommand{\url}[1]{\texttt{#1}}
\providecommand{\urlprefix}{URL }
\providecommand{\doi}[1]{https://doi.org/#1}

\bibitem{aliari2023face}
Aliari, M.A., Beauchamp, A., Popa, T., Paquette, E.: Face editing using part-based optimization of the latent space. In: CGF (2023)

\bibitem{bai2023ffhq}
Bai, H., Kang, D., Zhang, H., Pan, J., Bao, L.: Ffhq-uv: Normalized facial uv-texture dataset for 3d face reconstruction. In: CVPR (2023)

\bibitem{bai2021riggable}
Bai, Z., Cui, Z., Liu, X., Tan, P.: Riggable 3d face reconstruction via in-network optimization. In: CVPR (2021)

\bibitem{hifi3dface2021tencentailab}
Bao, L., Lin, X., Chen, Y., Zhang, H., Wang, S., Zhe, X., Kang, D., Huang, H., Jiang, X., Wang, J., Yu, D., Zhang, Z.: High-fidelity 3d digital human head creation from rgb-d selfies. TOG  (2021)

\bibitem{bao2021high}
Bao, L., Lin, X., Chen, Y., Zhang, H., Wang, S., Zhe, X., Kang, D., Huang, H., Jiang, X., Wang, J., et~al.: High-fidelity 3d digital human head creation from rgb-d selfies. TOG  (2021)

\bibitem{blanz2023morphable}
Blanz, V., Vetter, T.: A morphable model for the synthesis of 3d faces. In: SIGGRAPH (1999)

\bibitem{Bulat_2017_ICCV}
Bulat, A., Tzimiropoulos, G.: How far are we from solving the 2d \& 3d face alignment problem? (and a dataset of 230,000 3d facial landmarks). In: ICCV (2017)

\bibitem{chai2022realy}
Chai, Z., Zhang, H., Ren, J., Kang, D., Xu, Z., Zhe, X., Yuan, C., Bao, L.: Realy: Rethinking the evaluation of 3d face reconstruction. In: ECCV (2022)

\bibitem{chai2023hiface}
Chai, Z., Zhang, T., He, T., Tan, X., Baltrusaitis, T., Wu, H., Li, R., Zhao, S., Yuan, C., Bian, J.: Hiface: High-fidelity 3d face reconstruction by learning static and dynamic details. In: ICCV (2023)

\bibitem{cook1982reflectance}
Cook, R.L., Torrance, K.E.: A reflectance model for computer graphics. TOG  (1982)

\bibitem{dai2025high}
Dai, J., Wang, A., Ni, B., Cao, T.: High-quality facial albedo generation for 3d face reconstruction from a single image using a coarse-to-fine approach. arXiv:2506.13233  (2025)

\bibitem{danvevcek2022emoca}
Dan{\v{e}}{\v{c}}ek, R., Black, M.J., Bolkart, T.: Emoca: Emotion driven monocular face capture and animation. In: CVPR (2022)

\bibitem{EMOCA}
Danecek, R., Black, M.J., Bolkart, T.: Emoca: Emotion driven monocular face capture and animation. In: CVPR (2022)

\bibitem{deluigi2023inr2vec}
De~Luigi, L., Cardace, A., Spezialetti, R., Zama~Ramirez, P., Salti, S., Di~Stefano, L.: Deep learning on implicit neural representations of shapes. In: ICLR (2023)

\bibitem{debevec2012light}
Debevec, P.: The light stages and their applications to photoreal digital actors. In: ACM SIGGRAPH 2012 Courses. pp. 1--10 (2012)

\bibitem{deng2018uv}
Deng, J., Cheng, S., Xue, N., Zhou, Y., Zafeiriou, S.: Uv-gan: Adversarial facial uv map completion for pose-invariant face recognition. In: CVPR (2018)

\bibitem{deng2019accurate}
Deng, Y., Yang, J., Xu, S., Chen, D., Jia, Y., Tong, X.: Accurate 3d face reconstruction with weakly-supervised learning: From single image to image set. In: CVPRW (2019)

\bibitem{nextface}
Dib, A., Bharaj, G., Ahn, J., Thébault, C., Gosselin, P.H., Romeo, M., Chevallier, L.: Practical face reconstruction via differentiable ray tracing. EG  (2021)

\bibitem{dib2024mosar}
Dib, A., Hafemann, L.G., Got, E., Anderson, T., Fadaeinejad, A., Cruz, R.M., Carbonneau, M.A.: Mosar: Monocular semi-supervised model for avatar reconstruction using differentiable shading. In: CVPR (2024)

\bibitem{DECA}
Feng, Y., Feng, H., Black, M.J., Bolkart, T.: Learning an animatable detailed {3D} face model from in-the-wild images. In: SIGGRAPH (2021)

\bibitem{fitdiff}
Galanakis, S., Lattas, A., Moschoglou, S., Zafeiriou, S.: Fitdiff: Robust monocular 3d facial shape and reflectance estimation using diffusion models. In: WACV (2025)

\bibitem{gecer2021ostec}
Gecer, B., Deng, J., Zafeiriou, S.: Ostec: One-shot texture completion. In: CVPR (2021)

\bibitem{gecer2019ganfit}
Gecer, B., Ploumpis, S., Kotsia, I., Zafeiriou, S.: Ganfit: Generative adversarial network fitting for high fidelity 3d face reconstruction. In: CVPR (2019)

\bibitem{guo2020towards}
Guo, J., Zhu, X., Yang, Y., Yang, F., Lei, Z., Li, S.Z.: Towards fast, accurate and stable 3d dense face alignment. In: ECCV (2020)

\bibitem{hu2022lora}
Hu, E.J., Shen, Y., Wallis, P., Allen-Zhu, Z., Li, Y., Wang, S., Wang, L., Chen, W., et~al.: Lora: Low-rank adaptation of large language models. ICLR  (2022)

\bibitem{pix2pix2017}
Isola, P., Zhu, J.Y., Zhou, T., Efros, A.A.: Image-to-image translation with conditional adversarial networks. CVPR  (2017)

\bibitem{jiang2025distribution}
Jiang, D., Liu, D., Wang, Z., Wu, Q., Jin, X., Liu, D., Li, Z., Wang, M., Gao, P., Yang, H.: Distribution matching distillation meets reinforcement learning. arXiv:2511.13649  (2025)

\bibitem{kajiya1986rendering}
Kajiya, J.T.: The rendering equation. In: SIGGRAPH (1986)

\bibitem{karras2019style}
Karras, T., Laine, S., Aila, T.: A style-based generator architecture for generative adversarial networks. In: CVPR (2019)

\bibitem{karras2020analyzing}
Karras, T., Laine, S., Aittala, M., Hellsten, J., Lehtinen, J., Aila, T.: Analyzing and improving the image quality of stylegan. In: CVPR (2020)

\bibitem{labs2025flux1kontextflowmatching}
Labs, B.F., Batifol, S., Blattmann, A., Boesel, F., Consul, S., Diagne, C., Dockhorn, T., English, J., English, Z., Esser, P., Kulal, S., Lacey, K., Levi, Y., Li, C., Lorenz, D., Müller, J., Podell, D., Rombach, R., Saini, H., Sauer, A., Smith, L.: Flux.1 kontext: Flow matching for in-context image generation and editing in latent space. arXiv:2506.15742  (2025)

\bibitem{Lattas2023fitme}
Lattas, A., Moschoglou, S., Ploumpis, S., Gecer, B., Deng, J., Zafeiriou, S.: Fitme: Deep photorealistic 3d morphable model avatars. In: CVPR (2023)

\bibitem{lattas2020avatarme}
Lattas, A., Moschoglou, S., Gecer, B., Ploumpis, S., Triantafyllou, V., Ghosh, A., Zafeiriou, S.: Avatarme: Realistically renderable 3d facial reconstruction "in-the-wild". In: CVPR (2020)

\bibitem{Lattas2022AvatarMe++}
Lattas, A., Moschoglou, S., Ploumpis, S., Gecer, B., Ghosh, A., Zafeiriou, S.: Avatarme++: Facial shape and brdf inference with photorealistic rendering-aware gans. PAMI  (2022)

\bibitem{CelebAMask-HQ}
Lee, C.H., Liu, Z., Wu, L., Luo, P.: Maskgan: Towards diverse and interactive facial image manipulation. In: CVPR (2020)

\bibitem{lei2023hierarchical}
Lei, B., Ren, J., Feng, M., Cui, M., Xie, X.: A hierarchical representation network for accurate and detailed face reconstruction from in-the-wild images. In: CVPR (2023)

\bibitem{li2025lino}
Li, H., Chen, H., Ye, C., Chen, Z., Li, B., Xu, S., Guo, X., Liu, X., Wang, Y., Zhang, B., Ikehata, S., Shi, B., Rao, A., Zhao, H.: Light of normals: Unified feature representation for universal photometric stereo. arXiv:2506.18882  (2025)

\bibitem{li2024uv}
Li, H., Feng, Y., Xue, S., Liu, X., Zeng, B., Li, S., Liu, B., Liu, J., Han, S., Zhang, B.: Uv-idm: identity-conditioned latent diffusion model for face uv-texture generation. In: CVPR (2024)

\bibitem{li2025near}
Li, H., Ye, C., Chen, H., Xiao, W., Yan, Z., Xiao, L., Chen, Z., Xiang, J., Xu, S., Liu, X., Wang, Y., Zhang, B., Han, X., Yang, J., Zhao, H.: Near: Coupled neural asset-renderer stack. arXiv:2511.18600  (2025)

\bibitem{FaceKit}
Li, R., Bladin, K., Zhao, Y., Chinara, C., Ingraham, O., Xiang, P., Ren, X., Prasad, P., Kishore, B., Xing, J., Li, H.: Learning formation of physically-based face attributes. In: CVPR (2020)

\bibitem{FLAME}
Li, T., Bolkart, T., Black, M.J., Li, H., Romero, J.: Learning a model of facial shape and expression from {4D} scans. SIGGRAPH Asia  (2017)

\bibitem{adamw}
Loshchilov, I., Hutter, F.: Decoupled weight decay regularization. In: NeurIPS (2017)

\bibitem{lugaresi2019mediapipe}
Lugaresi, C., Tang, J., Nash, H., McClanahan, C., Uboweja, E., Hays, M., Zhang, F., Chang, C.L., Yong, M.G., Lee, J., et~al.: Mediapipe: A framework for building perception pipelines. arXiv:1906.08172  (2019)

\bibitem{Papantoniou2023Relightify}
Papantoniou, F., Lattas, A., Moschoglou, S., Zafeiriou, S.: Relightify: Relightable 3d faces from a single image via diffusion models. In: CVPR (2023)

\bibitem{paysan20093d}
Paysan, P., Knothe, R., Amberg, B., Romdhani, S., Vetter, T.: A 3d face model for pose and illumination invariant face recognition. In: AVSS (2009)

\bibitem{albedogan}
Rai, A., Gupta, H., Pandey, A., Carrasco, F.V., Takagi, S.J., Aubel, A., Kim, D., Prakash, A., De~la Torre, F.: Towards realistic generative 3d face models. In: WACV (2024)

\bibitem{ranjan2018generating}
Ranjan, A., Bolkart, T., Sanyal, S., Black, M.J.: Generating 3d faces using convolutional mesh autoencoders. In: ECCV (2018)

\bibitem{SMIRK}
Retsinas, G., Filntisis, P.P., Danecek, R., Abrevaya, V.F., Roussos, A., Bolkart, T., Maragos, P.: 3d facial expressions through analysis-by-neural-synthesis. In: CVPR (2024)

\bibitem{rombach2022high}
Rombach, R., Blattmann, A., Lorenz, D., Esser, P., Ommer, B.: High-resolution image synthesis with latent diffusion models. In: CVPR (2022)

\bibitem{RingNet}
Sanyal, S., Bolkart, T., Feng, H., Black, M.: Learning to regress 3d face shape and expression from an image without 3d supervision. In: CVPR (2019)

\bibitem{ADD}
Sauer, A., Lorenz, D., Blattmann, A., Rombach, R.: Adversarial diffusion distillation. In: ECCV (2024)

\bibitem{simeoni2025dinov3}
Sim{\'e}oni, O., Vo, H.V., Seitzer, M., Baldassarre, F., Oquab, M., Jose, C., Khalidov, V., Szafraniec, M., Yi, S., Ramamonjisoa, M., et~al.: Dinov3. arXiv:2508.10104  (2025)

\bibitem{smith2020morphable}
Smith, W.A.P., Seck, A., Dee, H., Tiddeman, B., Tenenbaum, J., Egger, B.: A morphable face albedo model. In: CVPR (2020)

\bibitem{Consistency}
Song, Y., Dhariwal, P., Chen, M., Sutskever, I.: Consistency models. In: ICML (2023)

\bibitem{LongCat-Image}
Team, M.L., Ma, H., Tan, H., Huang, J., Wu, J., He, J.Y., Gao, L., Xiao, S., Wei, X., Ma, X., Cai, X., Guan, Y., Hu, J.: Longcat-image technical report. arXiv:2512.07584  (2025)

\bibitem{team2025longcat}
Team, M.L., Ma, H., Tan, H., Huang, J., Wu, J., He, J.Y., Gao, L., Xiao, S., Wei, X., Ma, X., et~al.: Longcat-image technical report. arXiv:2512.07584  (2025)

\bibitem{team2025zimage}
Team, Z.I.: Z-image: An efficient image generation foundation model with single-stream diffusion transformer. arXiv:2511.22699  (2025)

\bibitem{tran2019towards}
Tran, L., Liu, F., Liu, X.: Towards high-fidelity nonlinear 3d face morphable model. In: CVPR (2019)

\bibitem{tran2018nonlinear}
Tran, L., Liu, X.: Nonlinear 3d face morphable model. In: CVPR (2018)

\bibitem{Wang_2025_CVPR}
Wang, C., Kang, D., Sun, H., Qian, S., Wang, Z., Bao, L., Zhang, S.H.: Mega: Hybrid mesh-gaussian head avatar for high-fidelity rendering and head editing. In: CVPR (2025)

\bibitem{wang2021review}
Wang, H.: A review of 3d face reconstruction from a single image. arXiv:2110.09299  (2021)

\bibitem{wang2022faceverse}
Wang, L., Chen, Z., Yu, T., Ma, C., Li, L., Liu, Y.: Faceverse: a fine-grained and detail-controllable 3d face morphable model from a hybrid dataset. In: CVPR (2022)

\bibitem{wang2021realesrgan}
Wang, X., Xie, L., Dong, C., Shan, Y.: Real-esrgan: Training real-world blind super-resolution with pure synthetic data. In: ICCVW (2021)

\bibitem{wang20243d}
Wang, Z., Zhu, X., Zhang, T., Wang, B., Lei, Z.: 3d face reconstruction with the geometric guidance of facial part segmentation. In: CVPR (2024)

\bibitem{woo2023convnext}
Woo, S., Debnath, S., Hu, R., Chen, X., Liu, Z., Kweon, I.S., Xie, S.: Convnext v2: Co-designing and scaling convnets with masked autoencoders. In: CVPR (2023)

\bibitem{wu2025qwenimagetechnicalreport}
Wu, C., Li, J., Zhou, J., Lin, J., Gao, K., Yan, K., ming Yin, S., Bai, S., Xu, X., Chen, Y., Chen, Y., Tang, Z., Zhang, Z., Wang, Z., Yang, A., Yu, B., Cheng, C., Liu, D., Li, D., Zhang, H., Meng, H., Wei, H., Ni, J., Chen, K., Cao, K., Peng, L., Qu, L., Wu, M., Wang, P., Yu, S., Wen, T., Feng, W., Xu, X., Wang, Y., Zhang, Y., Zhu, Y., Wu, Y., Cai, Y., Liu, Z.: Qwen-image technical report. arXiv:2508.02324  (2025)

\bibitem{wu2025fastavatar}
Wu, Y., Wu, Y., Li, W., Lu, Y., Feng, K., Chen, X.: Fastavatar: Towards unified fast high-fidelity 3d avatar reconstruction with large gaussian reconstruction transformers. arXiv:2508.19754  (2025)

\bibitem{yang2025freeuv}
Yang, X., Taketomi, T., Endo, Y., Kanamori, Y.: Freeuv: Ground-truth-free realistic facial uv texture recovery via cross-assembly inference strategy. In: CVPR (2025)

\bibitem{zheng2022imface}
Zheng, M., Yang, H., Huang, D., Chen, L.: Imface: A nonlinear 3d morphable face model with implicit neural representations. In: Proceedings of the IEEE/CVF conference on computer vision and pattern recognition. pp. 20343--20352 (2022)

\bibitem{ImFace++}
Zheng, M., Zhang, H., Yang, H., Chen, L., Huang, D.: Imface++: A sophisticated nonlinear 3d morphable face model with implicit neural representations. PAMI  (2025)

\bibitem{dml}
Zheng, Q., Deng, J., Zhu, Z., Li, Y., Zafeiriou, S.: Decoupled multi-task learning with cyclical self-regulation for face parsing. In: CVPR (2022)

\bibitem{zhou2024ultravatar}
Zhou, M., Hyder, R., Xuan, Z., Qi, G.: Ultravatar: A realistic animatable 3d avatar diffusion model with authenticity guided textures. In: CVPR (2024)

\bibitem{zielonka22mica}
Zielonka, W., Bolkart, T., Thies, J.: Towards metrical reconstruction of human faces. In: ECCV (2022)

\end{thebibliography}

\end{document}